%% file: main.tex
\documentclass[preprint,10pt,3p,twocolumn]{elsarticle}

\usepackage{amssymb}
\usepackage{url} 
\usepackage{amsmath}
\usepackage{rotating}
\usepackage{listings}
\usepackage{hyperref}
\usepackage{xcolor}
\usepackage{colortbl}
\definecolor{bgcolour}{rgb}{0.2,0.2,0.2}
\definecolor{green}{rgb}{0.6,0.8,0.2}
\definecolor{magenta}{rgb}{1.0,0.0,1.0}
\definecolor{codepurple}{rgb}{0.6,0.4,0.8}
\definecolor{gray}{rgb}{0.6,0.6,0.6}
\definecolor{codewhite}{rgb}{1.0,1.0,1.0}
\definecolor{promptcolor}{rgb}{0.0,0.6,0.0} 
\definecolor{magenta}{rgb}{1.0,0.0,1.0}
\definecolor{backcolour}{rgb}{0.95,0.95,0.92}
\definecolor{codegray}{rgb}{0.5,0.5,0.5}
\lstdefinestyle{command}{
    backgroundcolor=\color{bgcolour},
    commentstyle=\color{green},
    keywordstyle=\color{magenta},
    stringstyle=\color{codepurple},
    basicstyle=\ttfamily\footnotesize\color{codewhite},
    morekeywords={mycommand1, mycommand2, mycommand3},
    breakatwhitespace=false,
    breaklines=true,
    tabsize=4,
    numbersep=5pt
}

\begin{document}

\begin{frontmatter}



\title{Stereo Vision-Based Fall Prediction and Detection using Human Pose Estimation on the AMD Kria K26 SOM}


\author[sdsu]{Shreyas Narasimhiah Ramesh}
\author[psg]{P. D. Rathika}
\author[sdsu]{Mahasweta Sarkar}
\author[sdsu_psych]{Kristen Wells}
\author[odu]{Michel Audette}
\author[sdsu]{Christopher Paolini}

\affiliation[sdsu]{organization={Electrical and Computer Engineering, San Diego State University},
            addressline={5500 Campanile Drive}, 
            city={San Diego},
            postcode={92182-1309}, 
            state={California},
            country={USA}}

\affiliation[psg]{organization={Department of Robotics and Automation Engineering, PSG College of Technology},
            addressline={Peelamedu}, 
            city={Coimbatore},
            postcode={641004}, 
            state={Tamil Nadu},
            country={India}}
            
\affiliation[sdsu_psych]{organization={Psychology, San Diego State University},
            addressline={5500 Campanile Drive}, 
            city={San Diego},
            postcode={92182-4611}, 
            state={California},
            country={USA}}
            
\affiliation[odu]{organization={Electrical and Computer Engineering, Old Dominion University},
            addressline={5115 Hampton Blvd}, 
            city={Norfolk},
            postcode={23529}, 
            state={Virginia},
            country={USA}}

\begin{abstract}
  \input{1_abstract}
\end{abstract}



\twocolumn[
\begin{highlights}
  \setlength\itemsep{0em}
\item Design, implementation, and performance evaluation of a novel embedded, accelerated, vision-based fall detection system for elderly at-risk individuals.
\item Deployment is on an AMD System on a Module (SoM), with a Zynq UltraScale+  MultiProcessor System on Chip (MPSoC), which supports low power accelerated AI. 
\item Implementation employs a 3-stage pipeline architecture to process RGB and depth frames through a quantized YOLOX, A2J, and fall detection model.
\vspace{6in}
\end{highlights}
]

\end{frontmatter}

\section{Introduction}
\label{introduction}

\input{2_introduction}

\section{Related Works}
\label{relatedWorks}
\input{3_related_works}

\section{Methods}
\label{methodology}
\input{methods}
    




\section{Results}
\label{results}
\input{7_result}

\section{Discussion}
\label{discussion}
\input{9_discussion}

\section{Conclusion}
\label{conclusion}
\input{8_conclusion_and_future_works}

\section*{Acknowledgment}

Research reported in this publication/abstract was supported by the National Institute on Minority Health and Health Disparities of the National Institutes of Health under Award Numbers S21MD010690 (SDSU HealthLINK Endowment) and U54MD012397 (SDSU HealthLINK Center). The content is solely the responsibility of the authors and does not necessarily
represent the official views of the National Institutes of Health.

\section{CRediT authorship contribution statement}
\label{CRediT}
Conceptualization: Shreyas Narasimhiah Ramesh, Michel Audette, Christopher Paolini;
Methodology: Shreyas Narasimhiah Ramesh, Michel Audette, Christopher Paolini;
Software: Shreyas Narasimhiah Ramesh;
Validation: Shreyas Narasimhiah Ramesh;
Formal analysis: Shreyas Narasimhiah Ramesh;
Investigation: Shreyas Narasimhiah Ramesh, Christopher Paolini;
Resources: Shreyas Narasimhiah Ramesh, Mahasweta Sarkar, Kristen Wells, Christopher Paolini;Data Curation: Shreyas Narasimhiah Ramesh;
Writing — original draft: Shreyas Narasimhiah Ramesh;
Writing — review \& editing: Christopher Paolini;
Visualization: Shreyas Narasimhiah Ramesh;
Supervision: Christopher Paolini;
Project administration: Christopher Paolini, Kristen Wells;
Funding acquisition: Christopher Paolini, Kristen Wells;


\section{Data availability}
\label{dataAvailability}
\textbf{GitHub: }\href{https://github.com/cpaolini/falldetection.git}{https://github.com/cpaolini/falldetection}








\bibliographystyle{siammod}
\bibliography{references}

\end{document}

%% file: 1_abstract.tex
\textbf{Background and Objective:} Falls among elderly people can lead to serious injuries and significantly impact their quality of life. Timely detection and prediction of falls are crucial for preventing fall-related injuries and improving the well-being of older adults. 
In this research, we propose a portable, low-power battery-operated, vision-based fall prediction and detection system using a human pose estimation (HPE) model running on an edge computing device, the AMD\textsuperscript{\tiny\textregistered} Kira\textsuperscript{\tiny\texttrademark}~K26 System-on-Module (SOM). Our objective is to create a non-intrusive and efficient solution for detecting potential fall events in real time. \\
\textbf{Methods: }The system consists of an Intel\textsuperscript{\tiny\textregistered} RealSense\texttrademark~D455 range-sensing camera connected to a K26 SOM through USB. The camera captures synchronized RGB and depth frames of pixel dimensions $640\times 480\times 3$ $(H\times W\times C)$ and $640\times 480$, respectively, at a rate of 60 frames per second (FPS) for real-time processing. The K26 SOM employs a 3-stage pipeline architecture to process the frames through quantized YOLOX, A2J, and fall detection machine learning models.
The YOLOX model takes each RGB frame to produce a bounding box for each human in the frame. The system then discards the RGB frame, preserving privacy. The second model, Anchor-to-Joint (A2J), accepts a depth frame and produces 15 key points of joint information for each detected human. The final model is a convolutional neural network (CNN) that takes informative joint coordinates $(x,y,z)$ and predicts human fall activity. \textcolor{black}{The YOLOX model was trained on the CrowdHuman dataset for human detection, the A2J model was trained on ITOP, MP-3DHP, UR Fall Detection, and a custom SDSU PSG dataset for pose estimation, and the fall prediction CNN was trained on UR Fall Detection and SDSU PSG datasets for binary fall classification.}The system is implemented on two different hardware architectures: a single-core DPU with the maximum possible resource executing a single thread and a two-core DPU to run YOLOX and A2J independently on each core executed using multiple threads.\\
\textbf{Results: }\textcolor{black}{The quantized accuracy—evaluated using Intersection over Union (IoU) of 50\% for YOLOX, mean Average Precision (mAP) with a 10-cm rule for A2J, and classification accuracy (TP + TN)/(TP + TN + FP + FN) for the fall prediction CNN—is 74\%, 84.13\%, and 75.85\%, respectively. The single-threaded serial pipeline achieved a throughput of approximately 2.5 FPS, while the multi-threaded implementation achieved up to 4.5 FPS.}\\
\textbf{Conclusion: }\textcolor{black}{
This research demonstrates the feasibility of a privacy-preserving, real-time fall detection system achieving 4.5 FPS on an AMD Kria K26 edge device. The proposed system validates that on-device human pose estimation and fall classification can be performed efficiently without cloud dependency, supporting practical applications in elderly monitoring and assistive healthcare environments. Future work will focus on improving model accuracy and processing speed for large-scale deployment.}

%% file: 2_introduction.tex
In the realm of healthcare and assisted living, the timely detection and prediction of falls stand as pivotal challenges with far-reaching implications, particularly for the elderly demographic. Falls are the leading cause of injury-related death and hospitalization in seniors~\cite{Lafleur2025_FallsOlderAdults, NCOA2025_FallsFacts, CDC2024_ADPH}, potentially resulting in disabilities, depression, and decreased quality of life. More than two-thirds of older adults fall at least once a year, with those in assisted living or long-term care facilities falling more frequently~\cite{Albasha, SHAO20231708}.

\textcolor{black}{Demographic trends show a rapidly expanding older adult population, intensifying the demand for effective fall-prevention and monitoring technologies.}

\textcolor{black}{Longer life expectancy has led to rising rates of chronic and age-related health conditions, further straining healthcare systems and increasing the urgency of preventive care.}

Fall-related risks are further magnified by the physical and psychological consequences that follow such events. Falls are the leading cause of traumatic injuries in seniors, often contributing to sustained functional decline and reduced independence.

\textcolor{black}{Falls continue to be the leading cause of injury-related morbidity and mortality among seniors.}~\cite{bergen2016falls}

\textcolor{black}{In the United States alone, millions of fall events occur annually, resulting in tens of thousands of deaths.}

Beyond the immediate physical impacts, fall events frequently generate long-term complications that jeopardize well-being and quality of life.

\textcolor{black}{Severe outcomes such as hip fractures carry significant mortality, with nearly one-fifth of patients dying within a year.}

\textcolor{black}{They are also a major contributor to traumatic brain injuries and long-term functional decline.}~\cite{Abrahamsen2009, haentjens2010meta}.

\textcolor{black}{Delayed assistance after a fall, often referred to as the “long lie,” can lead to serious complications including dehydration, hypothermia, and muscle breakdown.}

The broader impact of falls extends beyond clinical outcomes. The economic burden is immense, and the psychological effects—including fear of falling, reduced activity, and accelerated decline—add further complexity to fall management.

\textcolor{black}{The economic burden is equally substantial, with fall-related healthcare expenditures exceeding \$50 billion each year.}~\cite{Florence2018CDC}

\textcolor{black}{Psychological impacts such as the fear of falling can trigger reduced activity, isolation, and worsening physical and mental health.}~\cite{RN71}

Timely intervention plays a pivotal role in mitigating the impact of fall-related injuries, underscoring the critical need for accurate and efficient fall detection and prediction systems.  
\textcolor{black}{These factors highlight the importance of timely intervention, reinforcing the need for accurate and efficient fall-detection systems.}




\subsection{Existing Technologies and Limitations}
\textcolor{black}{
Fall detection systems generally rely on wearable sensors or vision-based monitoring, but both categories present limitations that hinder dependable real-world deployment. Wearable devices often suffer from user compliance issues, discomfort, and limited battery life, reducing their reliability for continuous monitoring~\cite{Li2009,Li2017}. }

\textcolor{black}{
Vision-based systems provide richer contextual information but raise privacy concerns, particularly in sensitive environments such as bedrooms or assisted-living facilities~\cite{zheng2015survey}. Their performance is also affected by occlusions, lighting variations, and the fact that cameras only monitor the specific spaces in which they are installed~\cite{Kwolek2014,Huang2018}. }

\textcolor{black}{
Embedded and edge-based solutions must additionally operate under strict computational and power constraints, making real-time inference challenging without hardware acceleration~\cite{Torti2018}. These limitations highlight the need for fall detection approaches that are privacy-preserving, robust to environmental variations, and capable of efficient real-time execution on resource-constrained platforms.}

\subsection{Human Pose Estimation}
Human pose estimation (HPE), the process of identifying and tracking key points or joints on the human body from visual data, has emerged as a fundamental technology for fall detection and prediction. By accurately extracting human poses, researchers can gain insights into body movements and postures, enabling the detection of abnormal patterns indicative of potential falls. Human pose estimation offers a non-intrusive approach while preserving user privacy, making it a promising avenue for fall detection systems.~\cite{openpose, HRNet}

\textcolor{black}{Modern HPE systems build upon advancements in deep learning that enable precise joint localization in both 2D and 3D, making them suitable for analyzing critical activities of daily living and identifying deviations associated with fall events.}~\cite{a2j}

\subsection{Research Objectives}
The primary objective of this research is to develop an accurate and efficient fall detection and prediction system using human pose estimation techniques on the AMD Kria K26 SOM platform. Specific objectives include exploring state-of-the-art neural networks, designing a real-time pipeline, developing robust fall detection techniques, optimizing performance, conducting comprehensive evaluations, and exploring integration with other assistive technologies.

\textcolor{black}{These objectives directly address the limitations of conventional camera-based and wearable approaches by leveraging depth-based pose extraction and embedded AI acceleration to achieve privacy-preserving, real-time performance.}

\subsection{Proposed Approach}

\textcolor{green}{Existing wearable and camera-based fall detection systems suffer from issues such as user discomfort, privacy concerns, sensitivity to lighting and occlusions, and limited robustness in real-world environments.} 
\textcolor{black}{The proposed design addresses these limitations by relying on depth data and skeletal pose representations, which significantly reduce privacy intrusion compared to RGB video and remain resilient under varying illumination conditions.} 
\textcolor{black}{Human pose estimation further mitigates the effects of occlusions and background clutter by extracting structural body keypoints rather than operating directly on raw pixel values.} 
\textcolor{black}{Additionally, deploying the full inference pipeline on the AMD Kria K26 SOM enables efficient, on-device processing, eliminating cloud dependence, reducing latency, and ensuring reliability in resource-constrained settings.} 
\textcolor{black}{These features collectively provide a privacy-preserving, robust, and real-time fall detection solution that overcomes the fundamental challenges of prior approaches.} 

\subsection{Contributions}
The primary contribution of this research is the development of an embedded, hardware-accelerated, low-power, and portable fall prediction and detection system that utilizes human pose estimation techniques. Additionally, a dataset of human pose kinematics features from controlled fall experiments has been created, providing valuable insights for training and evaluating fall detection algorithms. The system is designed to be privacy-preserving, resource-aware, and robust, ensuring its effectiveness and practicality in real-world scenarios.

\textcolor{black}{Furthermore, this work bridges advances in edge intelligence and embedded AI by demonstrating an FPGA-accelerated fall detection pipeline capable of real-time performance on the AMD Kria K26 SOM platform. This positions the system as a practical and scalable solution for deployment in home-care and assisted-living environments.}

\textcolor{black}{The remainder of this paper is organized as follows: Section~2  provide an overview of prior research in sensor-based and vision-based fall detection; Section~3 describes the system architecture and methodology; Section~3 presents the dataset, training details, and experimental setup; Section~4 reports results and discusses model performance; and Section~5 concludes the paper with limitations and future work.}

%% file: 3_related_works.tex
Fall detection technologies have evolved considerably over the past two decades, particularly with the transition from threshold-based and binary decision systems to more sophisticated multi-class and severity-aware models. Early work largely relied on wearable sensors, such as accelerometers and gyroscopes, to capture motion signatures indicative of falls~\cite{Li2009}. These systems typically analyzed acceleration thresholds or angular velocity patterns, and later incorporated both sensing modalities to improve robustness and reduce false alarms. The integration of deep learning techniques, including Convolutional Neural Networks (CNNs), further enhanced classification accuracy in wearable fall detection systems~\cite{Li2017, Hwang2017}. More recent advancements explore the feasibility of implementing machine learning and fall-detection algorithms directly on microcontroller units (MCUs) and edge devices, supported by the increasing computational capability of embedded platforms~\cite{Mrozek2020, ElAttaoui2020}.

In parallel with MCU-based approaches, hardware-accelerated and FPGA-based fall detection systems have emerged as a promising class of edge-intelligent solutions. Tian et al.~\cite{Tian2024_SmartHealth} introduced an ultra low-power, wearable architecture that accelerates shallow-learning fall detection models for elderly at-risk individuals. Their subsequent work demonstrated efficient fall detection through edge inferencing on embedded hardware platforms~\cite{Tian2024_ISICN}. Complementary research by Bharathkumar et al.~\cite{Bharathkumar2020_FPGAFall} implemented FPGA-based edge inferencing for fall detection, highlighting the benefits of hardware acceleration in achieving low-latency and energy-efficient operation. These developments illustrate a broader shift toward deploying fall detection models on specialized hardware to improve responsiveness and reduce computational load.

Vision-based fall detection approaches have also gained significant traction due to their non-intrusive nature and ability to capture rich contextual information~\cite{Espinosa2019, Bevilacqua2014}. Traditional RGB methods leveraged background subtraction, object tracking, and motion analysis to identify fall-related patterns~\cite{Huang2018}. Despite their effectiveness, these methods are often sensitive to illumination changes, occlusions, and differences in clothing or body appearance. Depth-based systems, enabled by devices such as the Microsoft Kinect, provided a more robust alternative by capturing three-dimensional structural information and using it to estimate body pose and movement trajectories~\cite{Solbach2017}. Recent work has focused on extracting 3D human pose, motion dynamics, and postural transitions to detect falls with higher reliability.

To address privacy concerns associated with RGB imaging, researchers have experimented with obfuscated or anonymized visual representations, such as silhouette extraction, thermal imaging, or depth-only video streams~\cite{Liu2020, pmlr-v116-asif20a}. These privacy-preserving modalities help mitigate ethical and user-acceptance challenges, particularly in sensitive environments such as assisted living facilities.

Despite significant progress, several challenges remain. Many existing systems struggle with occlusions, cluttered backgrounds, and multi-person environments, reducing their reliability in real-world deployments~\cite{Rajagopalan2017, Igual2013}. Additionally, computation-heavy deep learning architectures hinder real-time performance on resource-constrained platforms, motivating research into model optimization, hardware acceleration, and embedded processing~\cite{Torti2018, Chen2021, Bagal2012}. Deployment scalability also remains a challenge, as vision-based systems often require environment-specific calibration and may fail to generalize across different architectural layouts or user behaviors~\cite{Zhang2015, Igual2013}.

%% file: methods.tex
\subsection{System Overview}
\begin{figure*}[htbp]
    \centering
    \includegraphics[width=1\textwidth]{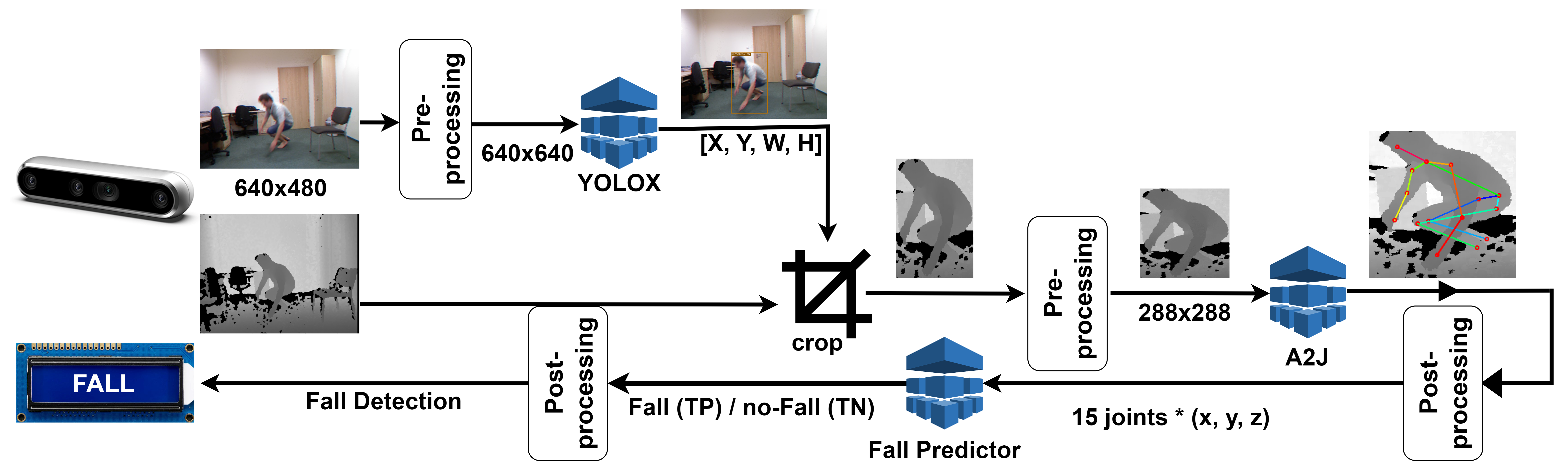}
    \caption{The proposed fall detection method. First, a human subject in a frame is detected using YOLOX~\cite{yolox} Second, the A2J model~\cite{a2j} is applied on the depth frame to estimate the joint pose information. Third, a CNN model analyses a set of frames' joint information to detect the occurrence of fall events. Images are taken from the UR Fall Detection dataset~\cite{urfall}.}
    \label{fig:system_overview}
\end{figure*}
Figure \ref{fig:system_overview} shows the overview of the proposed fall prediction and detection system. The system is comprised of an Intel\textsuperscript{\textregistered} RealSense\textsuperscript{\tiny\texttrademark} D455 range-sensing camera~\cite{realsense} connected to a AMD\textsuperscript{\tiny\textregistered} Kria\textsuperscript{\tiny\texttrademark} KV260 Vision AI~\cite{githubVitisAI} development platform. The camera produces frames at a resolution of $640 \times 480$ pixels and a frame rate of 60 frames per second (fps), allowing the system to process and operate in real time. RGBD frames from the camera are processed through a combination of machine learning models on the KV260 to predict the event of a human fall. \\
Upon the availability of a synchronized frameset, the RGB frame is first extracted and provided as input to the YOLOX model~\cite{yolox} for human detection. This model processes the full RGB image and returns bounding box coordinates delineating any detected human subjects in the pixel space, taking the form [\textit{xmin, ymin, width, height}] within the frame dimensions. With humans localized in the scene, the depth frame is subsequently extracted from the frameset and preprocessed to isolate the region of interest using the estimated boundary box. \textcolor{blue}{If there is no person identified int he frame of view then the RGB and depth frame pair is dropped.} The cropped depth frame is then passed to the A2J~\cite{a2j} pose estimation model which outputs a list of 15 joint coordinates for each detected human in 3D space as [\textit{x, y, z}] triplets. Finally, the predicted joint positions are analyzed by the fall prediction model to categorize the pose as either a fall event, or normal posture, through binary classification~\cite{10388590}.
\subsection{Data Acquisition}
The data acquisition process involves capturing synchronized RGB and depth frames using an Intel\textsuperscript{\tiny\textregistered} RealSense\textsuperscript{\tiny\texttrademark} D455 depth camera. This camera employs active infrared stereo vision technology to generate depth information, enabling the system to perceive the 3D spatial characteristics of the scene. The depth camera captures frames at a resolution of $640 \times 480$ pixels and a frame rate of 60 frames per second (fps), facilitating real-time processing.
The librealsense SDK \cite{realsenseSDK} is employed to interface with the depth camera, enabling seamless color and depth video streaming, as well as providing intrinsic and extrinsic calibration information. This SDK also offers synthetic streams, such as pointclouds and depth alignment to color frames, along with built-in support for recording and replaying streaming sessions in the ROSBag file format.

\subsection{Kria KV260 Vision AI Platform}
The AMD\textsuperscript{\tiny\textregistered}~Kria\textsuperscript{\tiny\texttrademark}~KV260 Vision AI development platform is a powerful and efficient edge computing solution designed for accelerating deep learning and computer vision applications. At the heart of this platform lies the Xilinx Kria K26 System-on-Module (SOM), which combines a quad-core Arm Cortex-A53 CPU with a dual-core Arm Cortex-R5 real-time processor and a dedicated deep learning processing unit (DPU).

The DPU, based on the Xilinx Deep Learning Processor (DLP) architecture, provides hardware acceleration for convolutional neural networks (CNNs), enabling high-performance and energy-efficient inference at the edge. The Kria KV260 platform is well-suited for applications requiring real-time processing of visual data, making it an ideal choice for the proposed fall prediction and detection system.

\subsection{Machine Learning Models}
The fall detection system leverages three main machine learning models: YOLOX for human detection, Anchor-to-Joint Regression Network (A2J) for human pose estimation\cite{a2j}, and a Convolutional Neural Network (CNN) for fall detection and classification. These models are briefly summarized as follows:

\subsubsection{YOLOX}
The primary goal is to detect humans within the camera's field of view, achieved through the utilization of the YOLOX-S object detection model. The YOLOX model is an anchor-free approach known for its exceptional speed and accuracy, requiring fewer operations during inference compared to anchor-based models like YOLOv3~\cite{yolov3} and YOLOv4~\cite{yolov4}. 
Several modifications were made to the model architecture: The activation function was changed from MISH to LeakyReLU, the maxpool sizes of the spatial pyramid pooling (SPP) were changed to 3$\times$3, 5$\times$5, and 7$\times$7, as the DPU has a maximum kernel size of 8$\times$8 as the limitation. Lastly, the number of output labels was limited to only one, i.e., a single person. The YOLOX model was trained on the \textit{CrowdHuman} dataset~\cite{crowdhuman}, a large dataset of human detection images. The primary evaluation metric for object detection is Average Precision (AP) for the person class, representing how accurately the model detects persons. Intersection over Union (IoU) is a key component in calculating AP. It measures the overlap between the predicted bounding box $B_p$ and ground truth box $B_{gt}$:
\begin{figure}[htbp]
\centering
\includegraphics[width=0.1\textwidth]{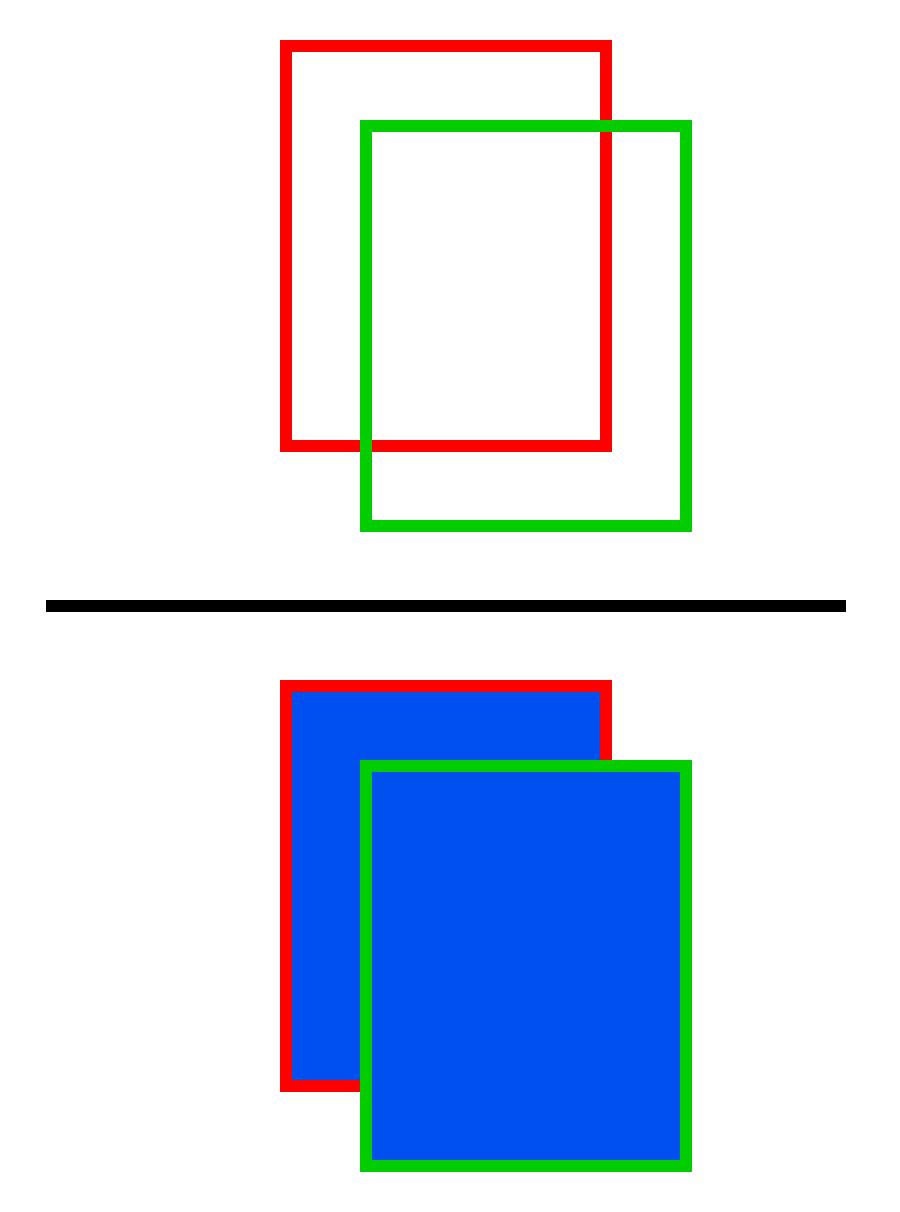}
\caption{Intersection of Union Concept}
\label{fig:iou}
\end{figure}
$$IoU = \frac{\text{Area of Intersection } (B_p \cap B_{gt})}{\text{Area of Union } (B_p \cup B_{gt})}$$
A prediction is considered a true positive if its IoU with the ground truth exceeds a threshold like 0.5. The IoU directly indicates localization accuracy for object detection. Maximizing IoU typically improves mAP as higher overlap with ground truth equates to better detection performance. The model produces the boundary box coordinates in the format [\textit{xmin, ymin, width, height}], where \textit{xmin, ymin} are the left-top corner of the bounding box. The modified YOLOX model is able to achieve good accuracy on real-time streams from the depth camera.

\subsubsection{Human Pose Estimation}
Accurate human pose estimation is a crucial component of the fall prediction and detection system. This work leverages the Anchor-to-Joint Regression Network (A2J) \cite{a2j}, a deep learning model that infers 3D human body pose from a single depth image. The A2J model employs a 2D CNN ResNet-50 architecture pretrained on the ImageNet dataset \cite{ILSVRC15}~as the backbone network, enabling efficient operation on the resource-constrained Kria K26 SOM platform.
\begin{figure}[htbp]
    \centering
    \includegraphics[width=0.25\textwidth]{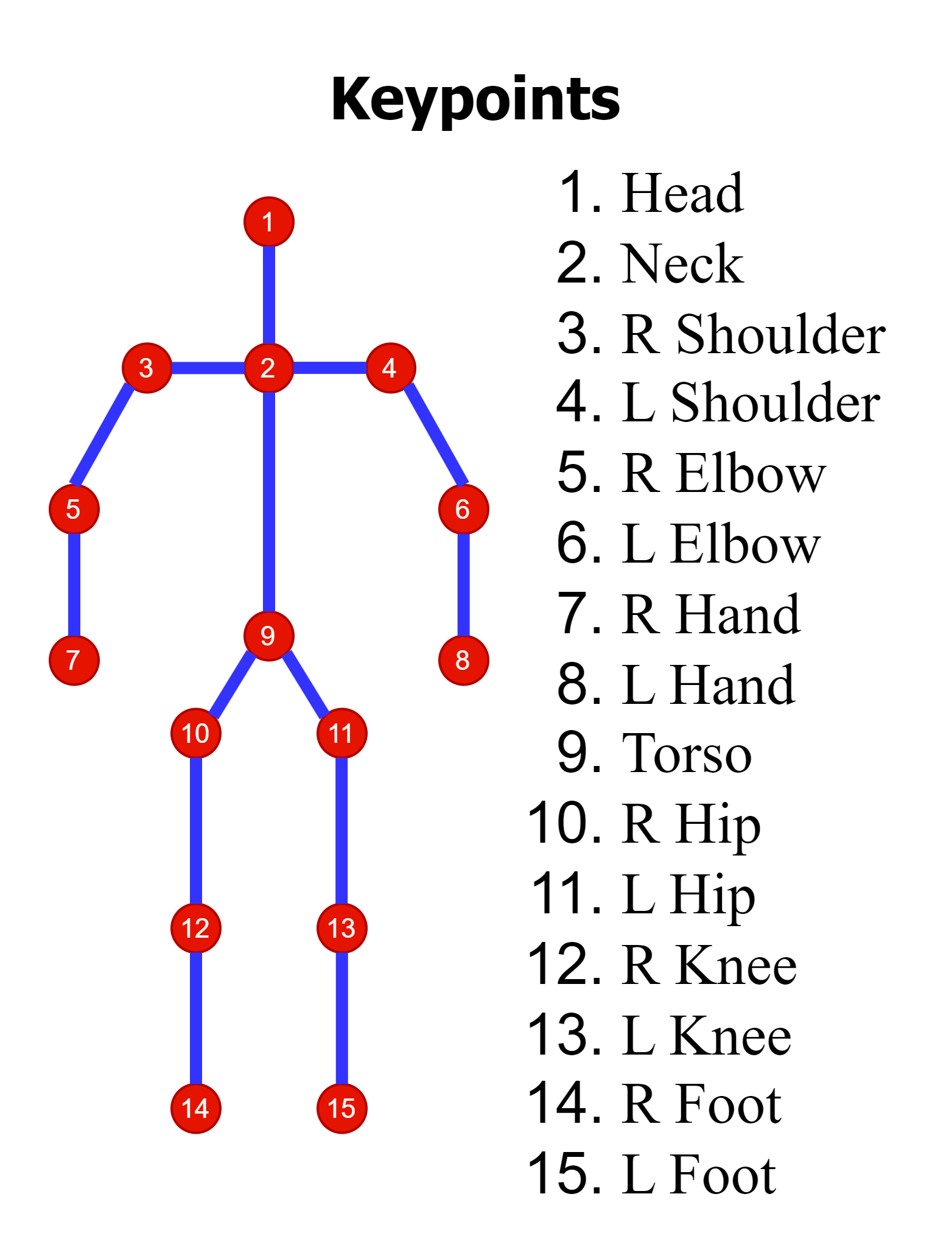}
    \caption{Full human body pose joints of interest.}
    \label{fig:skeleton}
\end{figure}
The A2J model is trained on multiple datasets, including ITOP \cite{itop}, MP-3DHP \cite{mp3dhp}, UR Fall Detection \cite{urfall}, and a custom SDSU\textunderscore PSG dataset. The model predicts the positions of 15 key joints in the human body, represented as 3D coordinates in the camera space.
\subsubsection{Fall Detection and Classification}
The fall detection and classification component of the system utilizes a Convolutional Neural Network (CNN) model \cite{fallDetectorModel} to analyze the predicted 3D human pose information from the A2J model. This CNN model is trained on the UR Fall Detection Dataset \cite{urfall} and the custom SDSU\textunderscore PSG dataset, comprising sequences of falls and activities of daily living (ADL).
The CNN model processes the temporal sequence of 3D joint positions and classifies each frame as either a "fall" or "non-fall" event. The model's performance is evaluated using a confusion matrix, enabling the calculation of essential metrics such as precision, recall, and F1 score, providing a comprehensive assessment of its accuracy in distinguishing between fall and non-fall events.

\subsection{Model Quantization and Deployment}
\begin{figure}[htbp]
    \centering
    \includegraphics[width=0.47\textwidth]{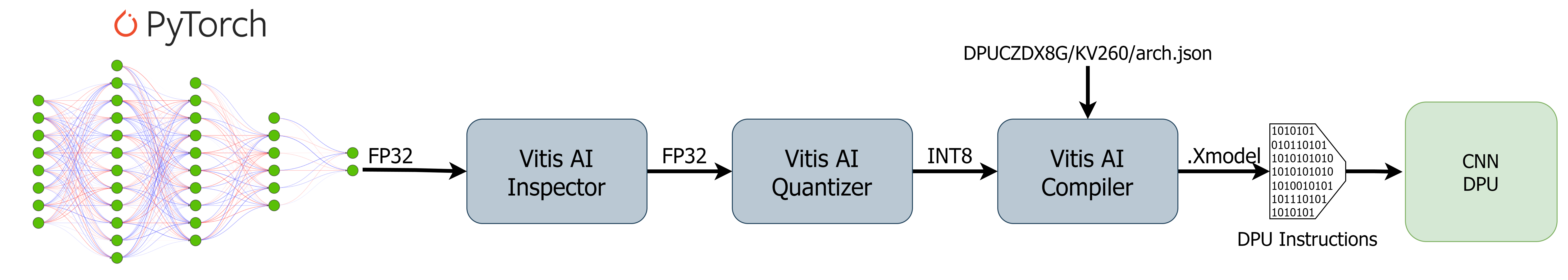}
    \caption{Steps involved in the model development of the pre-trained FP32 PyTorch\texttrademark~AI models into INT8 for inference on the Kria\texttrademark~K26 SOM.}
    \label{fig:deploymentSteps}
\end{figure}
The Vitis AI framework from AMD is leveraged to optimize and deploy the deep learning models on the Kria KV260 platform, taking advantage of its hardware acceleration capabilities. The key components of the Vitis AI stack~\cite{vitisai} involved in system optimization workflow are:
\begin{enumerate}
    \item Model Inspector: Before quantizing the pre-trained float model, it is examined with the Vitis AI Model Inspector, which returns partition information showing which operators may operate on which device (DPU/CPU).
    \item Model Quantization: The Vitis AI Quantizer lowers processing complexity without sacrificing prediction accuracy by converting 32-bit floating-point (FP32) weights and activations to 8-bit fixed-point integers (INT8). The pre-trained FP32 PyTorch models (YOLOX, A2J, and Fall Predictor CNN) are quantized to INT8 precision using the Vitis AI Quantizer tool.
    \item Model Compilation: The quantized INT8 models are compiled using the Vitis AI Compiler, which generates an optimized executable model (xmodel file) tailored for efficient execution on the Kria K26 SOM's DPU hardware.
    \item Application Integration: The compiled xmodel is integrated into the software application running on the PetaLinux operating system using the VART (Vitis AI RunTime) APIs. This includes deserializing the xmodel, creating DPU runners, managing input/output tensors, and asynchronous job execution on the DPU.
    \item Runtime Optimization: During inference, the application leverages the VART APIs to optimally utilize the DPU hardware acceleration capabilities. This includes techniques like asynchronous job submission, multi-threading, and efficient data transfer between the CPU and DPU.
\end{enumerate}

\subsection{Deep Learning Processor Unit}
The AMD\textsuperscript{\textregistered} Kria\textsuperscript{\tiny\texttrademark} K26 SOM features a Deep Learning Processing Unit (DPU), a customizable compute engine tailored for convolutional neural networks. The \verb|DPUCZDX8G|, designed for the UltraScale\small{\texttt{+}} MPSoC, seamlessly integrates into the programmable logic (PL) and establishes direct connectivity to the processing system (PS) through an AXI interconnect \cite{dpuProductGuide}. Upon reset, instructions are fetched from off-chip memory to govern the computing engine's operations, with these instructions generated by the Vitis AI compiler for optimized model layer performance.

\subsubsection{Integrating the DPU into Custom Platforms}
Two primary workflows are supported for integrating the DPUCZDX8G: the Vivado flow and the Vitis flow \cite{dpuWorkflow}. In this research, the Vivado flow was chosen, which involves creating a hardware platform project in Vivado, configuring the hardware for an accelerated design, synthesizing, and generating a bitstream file.

\subsubsection{Possible DPU Architectures on the KV260}
The AMD\textsuperscript{\textregistered} Kria\textsuperscript{\tiny\texttrademark}~K26 SOM's programmable logic offers various logical resources, including 256K system logic cells, 64 UltraRAM blocks, 144 block RAM blocks, and 1.2K DSP slices. These resources enable the synthesis of different DPU architectures within the programmable logic fabric. The available DPU configurations range from a single high-performance core to multiple lower-performance cores, catering to diverse computational requirements. For instance, a single core B4096 architecture can be implemented, capable of executing 4096 multiply-accumulate (MAC) operations per clock cycle at a frequency of 325 MHz. Alternatively, multiple homogeneous cores can be instantiated, such as two B1600 cores or three B1024 cores, enabling parallel execution of multiple models or workloads.

\subsubsection{Chosen DPU Architecture}
To achieve optimal system throughput, three distinct hardware architectures were explored and evaluated:
\begin{enumerate}
    \item \textbf{Single-core B4096 DPU design:}
The single-core B4096 architecture allows for the sequential execution of AI models on the DPU. With the maximum available logical resources, this architecture executes 4096 multiply-accumulate (MAC) operations per clock cycle at 325 MHz, providing a higher inference time per model execution. The throughput can be further boosted by employing data parallelism and multi-threading techniques for the DPU runners.
While the single-core design has limited parallel processing capacity, which may impact throughput in scenarios demanding simultaneous model execution, its higher performance per core and the ability to leverage data parallelism and multi-threading make it a compelling choice for our application.
    \item \textbf{Dual-core B1600 DPU design:}
The dual-core B1600 architecture features two homogeneous B1600 DPU cores connected to the Zynq UltraScale+ MPSoC processing system (PS) through an AXI interconnect. This design allows for the parallel execution of models, with YOLOX and A2J running on dedicated DPU cores, while the fall prediction model is executed on the CPU itself.
The dual-core approach offers enhanced throughput by processing input data concurrently on separate cores. However, it introduces potential challenges in balancing the workload between cores and introduces overhead due to offloading the fall prediction model to the CPU.
    \item \textbf{Three-core B1024 DPU design:}
The three-core B1024 architecture allocates separate B1024 DPU cores to each of the three models: YOLOX, A2J, and fall prediction. This design enables parallel execution of all three models, potentially improving overall throughput.
However, the limitations in the operations capability of the B1024 architecture DPU core lead to poor inference results for the computationally intensive YOLOX model. Achieving a balance between model complexity and the available core resources becomes crucial in this scenario.
\end{enumerate}

\subsubsection{Resource Utilization}
\begin{figure}[htbp]
    \centering
    \includegraphics[width=0.4\textwidth]{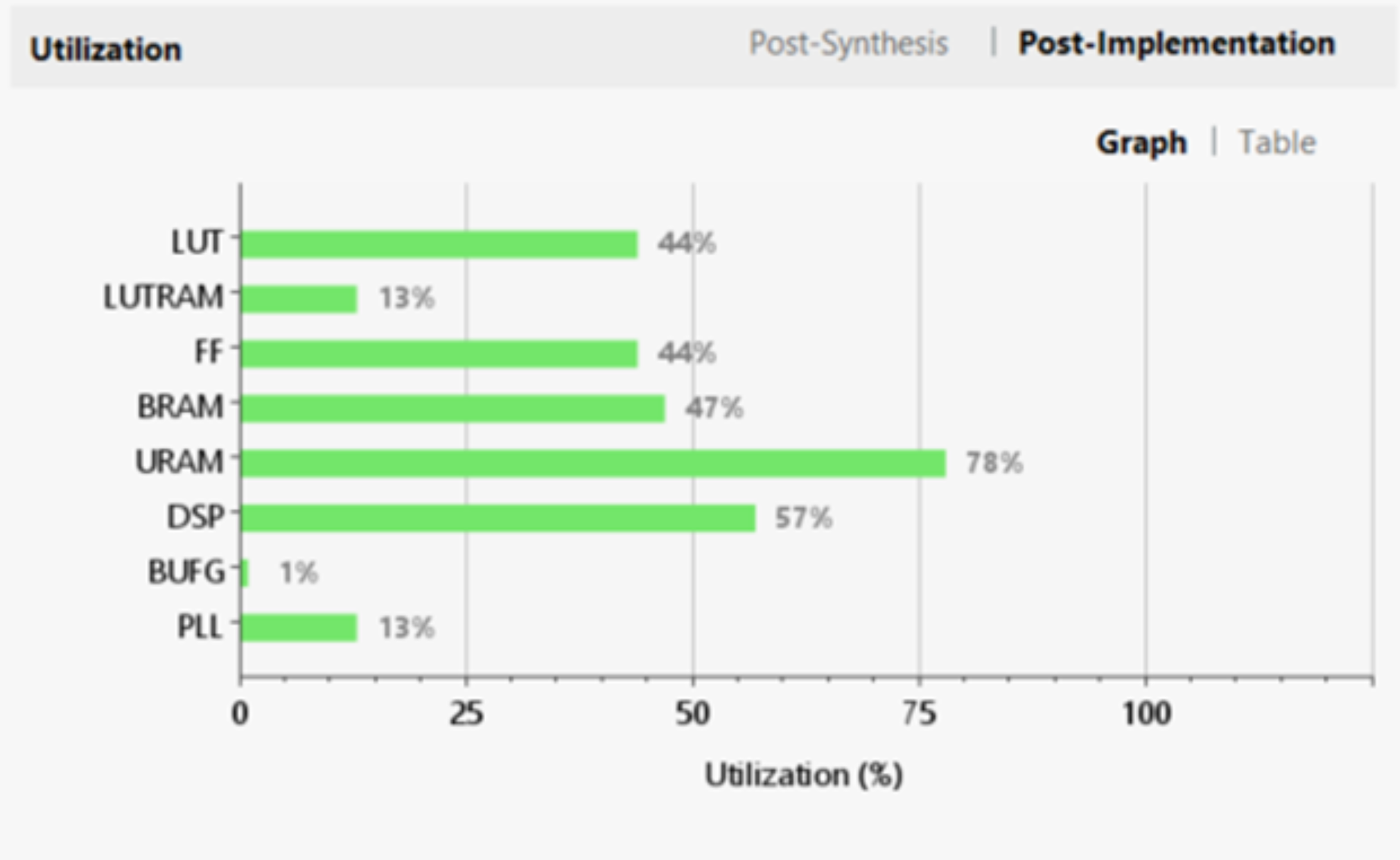}
    \caption{FPGA Resource Utilization of the 1 core B4096 DPU Design.}
    \label{fig:resourceUtilization_4096}
\end{figure}
\begin{figure}[htbp]
    \centering
    \includegraphics[width=0.4\textwidth]{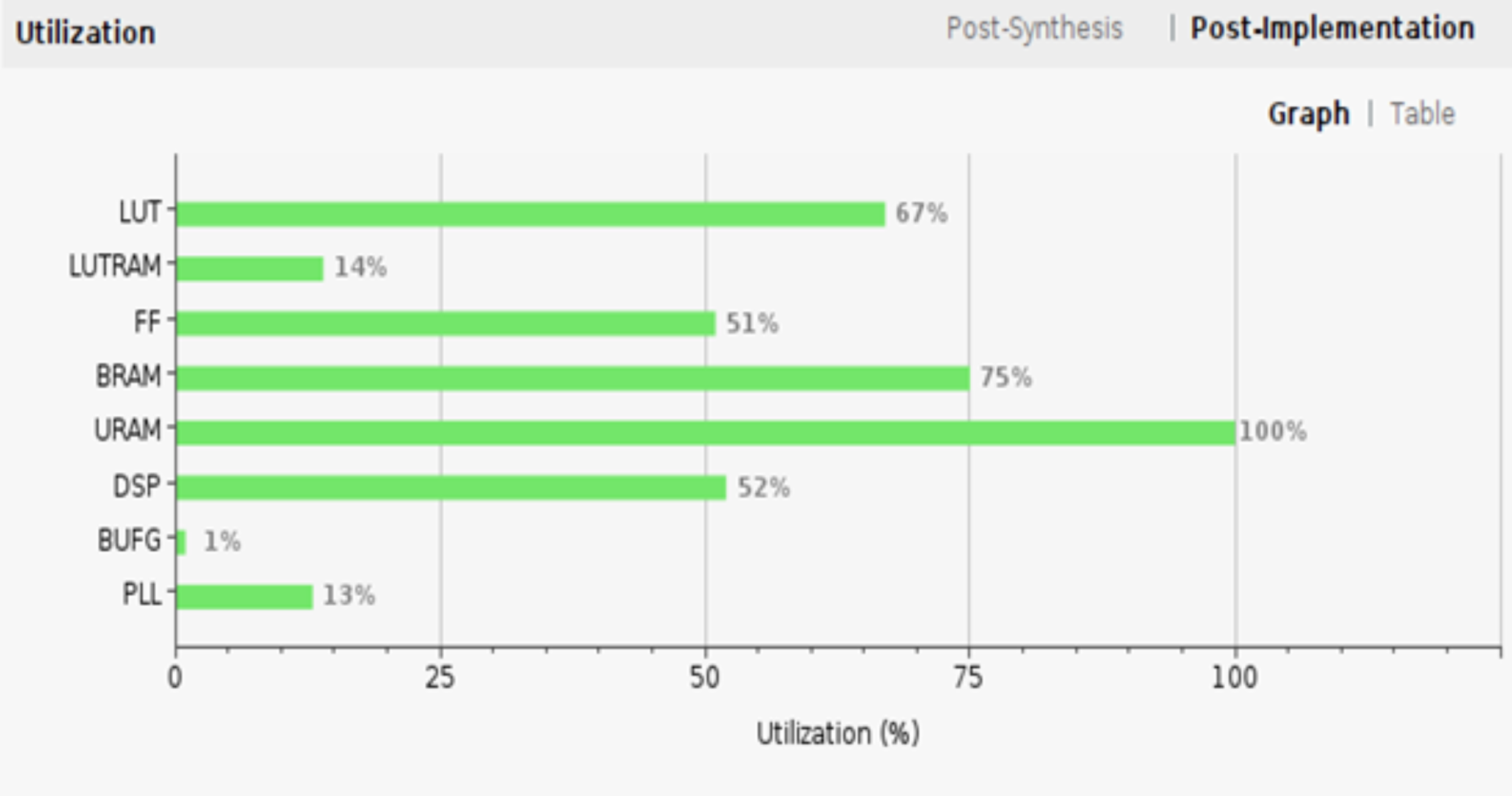}
    \caption{FPGA Resource Utilization of 2 core B1600 DPU Design.}
    \label{fig:resourceUtilization_1600}
\end{figure}
\begin{figure}[htbp]
    \centering
    \includegraphics[width=0.4\textwidth]{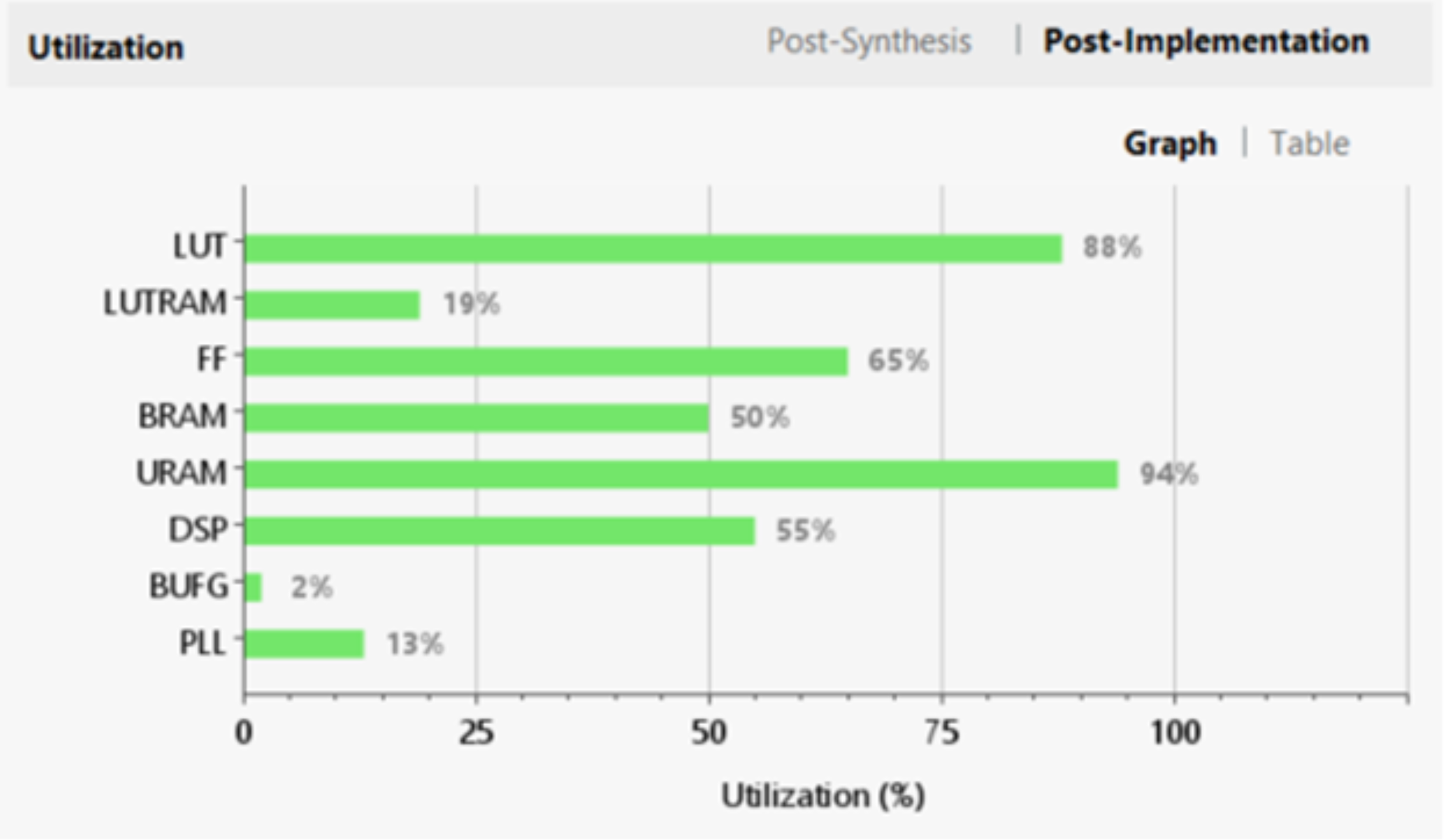}
    \caption{FPGA Resource Utilization of the 3 core B1024 DPU Design.}
    \label{fig:resourceUtilization_1024}
\end{figure}
Figures \ref{fig:resourceUtilization_4096}, \ref{fig:resourceUtilization_1600}, and \ref{fig:resourceUtilization_1024} summarize the logical resource utilization for the different DPU architectures explored.

\subsection{Embedded Linux, PetaLinux}
PetaLinux is an embedded Linux Software Development Kit (SDK) targeting
FPGA based system-on-chip (SoC) designs. The PetaLinux SDK contains everything
necessary to build, develop, test, and deploy embedded Linux systems. The PetaLinux
Board Support Package for the KV260 Starter Kit is used to build the embedded Linux to run on the KV260. Kernel is configured to enable the DPU device driver. Root filesystem is configured to include Vitis AI librarries and other supporting library packages within the build image. PetaLinux includes U-Boot, Linux kernel, Device Tree, and Root Filesystem components.

\subsubsection{Device Tree}
The device tree is a data structure used to describe the hardware components and their connections in a system. Using the XSCT (Xilinx Software Command-line Tool) command \verb|createdts| a device tree domain is created based on the hardware description specified by the \verb|XSA| file (hardware design). The \verb|-overlay| option is used to generate an overlay, which allows for dynamic device tree modifications without rebuilding the entire tree.

\subsubsection{Accelerated Application}
\begin{figure*}[htbp]
    \centering
    \includegraphics[width=1\textwidth]{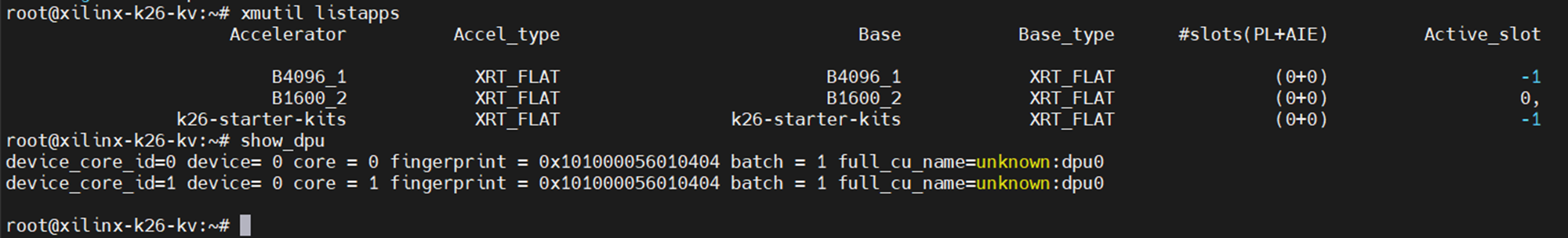}
    \caption{List of accelerated application available for execution.}
    \label{fig:acceleratedApplication}
\end{figure*}
To dynamically enable the accelerated applications on the AMD Kria K26 SOM we use the platform management utility called \verb|xmutil|. Figure \ref{fig:acceleratedApplication} shows the list of available applications

\subsection{Software Architecture}
\begin{figure}[htbp]
    \centering
    \includegraphics[width=0.47\textwidth]{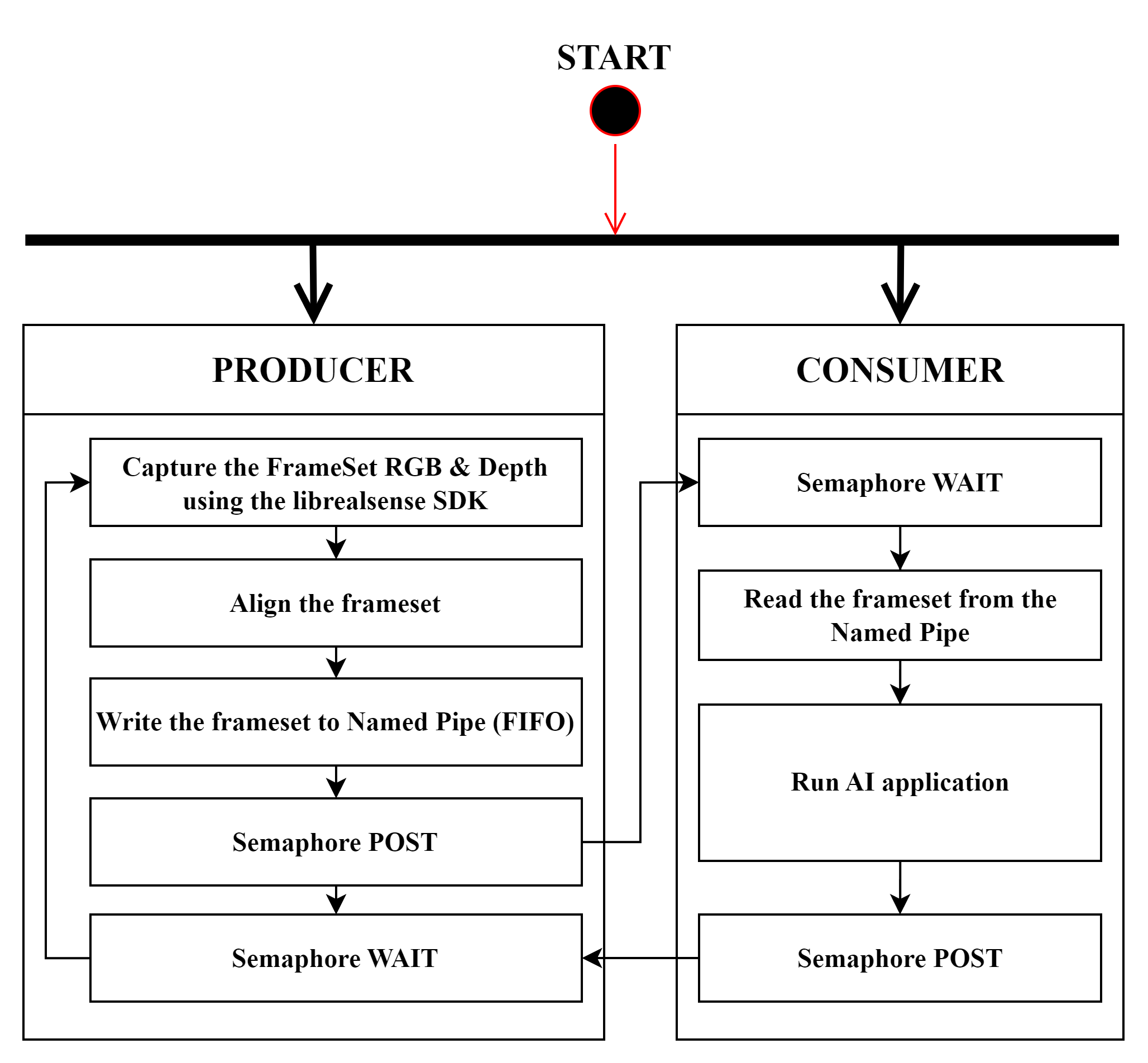}
    \caption{Producer-Consumer Software Architecture}
    \label{fig:prodCons}
\end{figure}
The proposed fall prediction and detection system is implemented using a producer-consumer software architecture, as illustrated in Figure \ref{fig:prodCons}. This architecture decouples the data acquisition process (producer) from the data processing and analysis tasks (consumer), enabling better concurrency, scalability, and fault tolerance.

\subsubsection{Data Acquisition and Preprocessing}
The producer process is responsible for acquiring data from either a pre-recorded video file or directly from the Intel RealSense D455 camera. The RealSense SDK \cite{realsense} is installed on the AMD Kria KV260 Vision AI Development Platform running PetaLinux, enabling seamless integration with the depth camera. The producer reads synchronized RGB and depth frames at a resolution of $640\times 480$ pixels and a frame rate of 60 fps, facilitating real-time processing.
To ensure data integrity during inter-process communication, header and trailer bytes are used as decorators to identify the start and end of frame data. The producer writes the frame data to a named pipe (FIFO) with a resized buffer to accommodate the large frame size. Semaphores are employed for synchronization, preventing race conditions and ensuring the integrity of data transfer between the producer and consumer processes.

\subsubsection{Serial Pipeline Implementation} \label{sec:serial-pipeline}
\begin{figure}[htbp]
    \centering
    \includegraphics[width=0.47\textwidth]{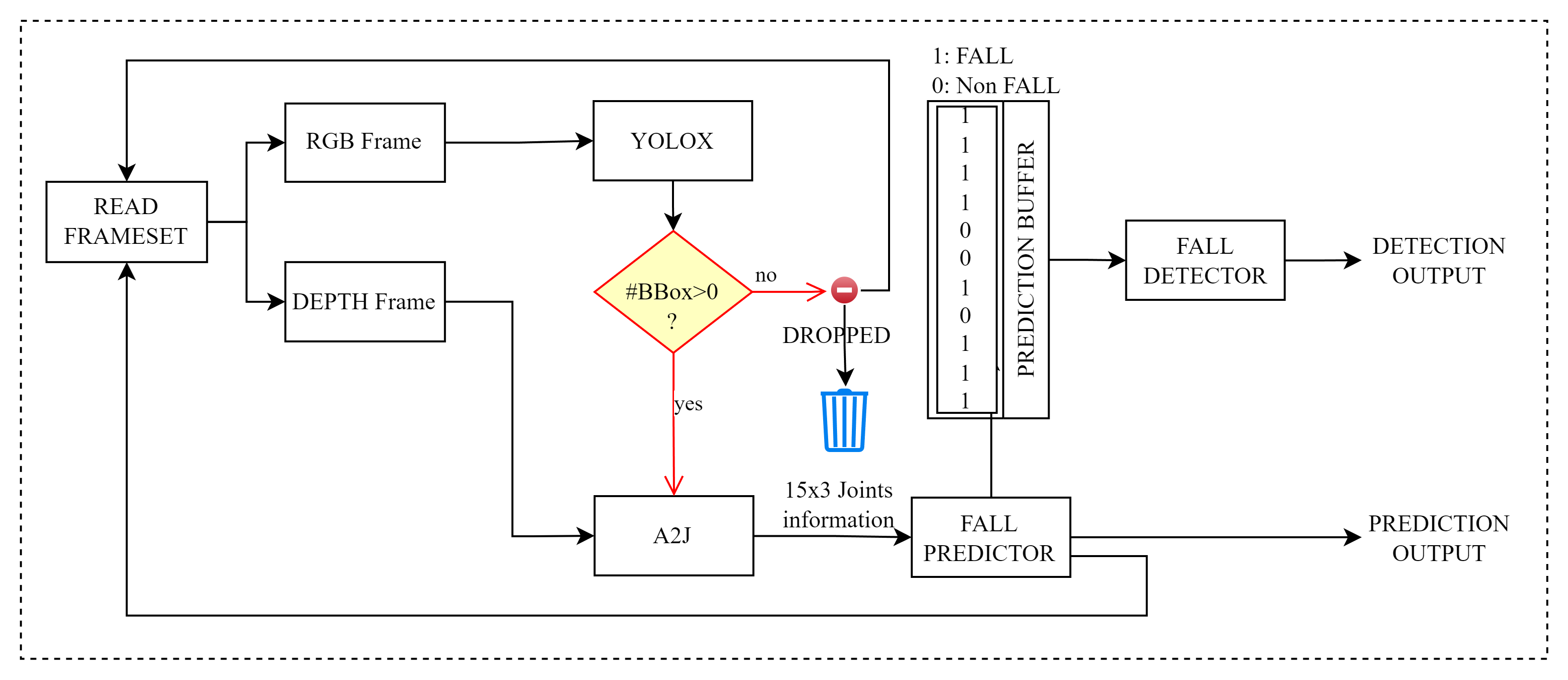}
    \caption{Serial Design Implementation.}
    \label{fig:serialPipeline}
\end{figure}
The serial pipeline design, as depicted in Figure \ref{fig:serialPipeline}, processes data sequentially using a single-core B4096 Deep Learning Processing Unit (DPU) architecture. The consumer process follows these steps:
\begin{enumerate}
    \item Read frame data from the FIFO pipe.
    \item Extract color and depth frames from the frame buffer.
    \item Run the YOLOX model \cite{yolox} for human detection on the color frame.
    \item If a person is detected, run the A2J model \cite{a2j} for pose estimation on the depth frame using the bounding box coordinates from YOLOX.
    \item Run the fall detection model \cite{fallDetectorModel} on the pose estimation output to classify the pose as a fall or non-fall event.
\end{enumerate}

\subsubsection{Parallel Pipeline Implementation} \label{sec:parallel-pipeline}
\begin{sidewaysfigure}[htbp]
    \centering
    \includegraphics[width=\textwidth]{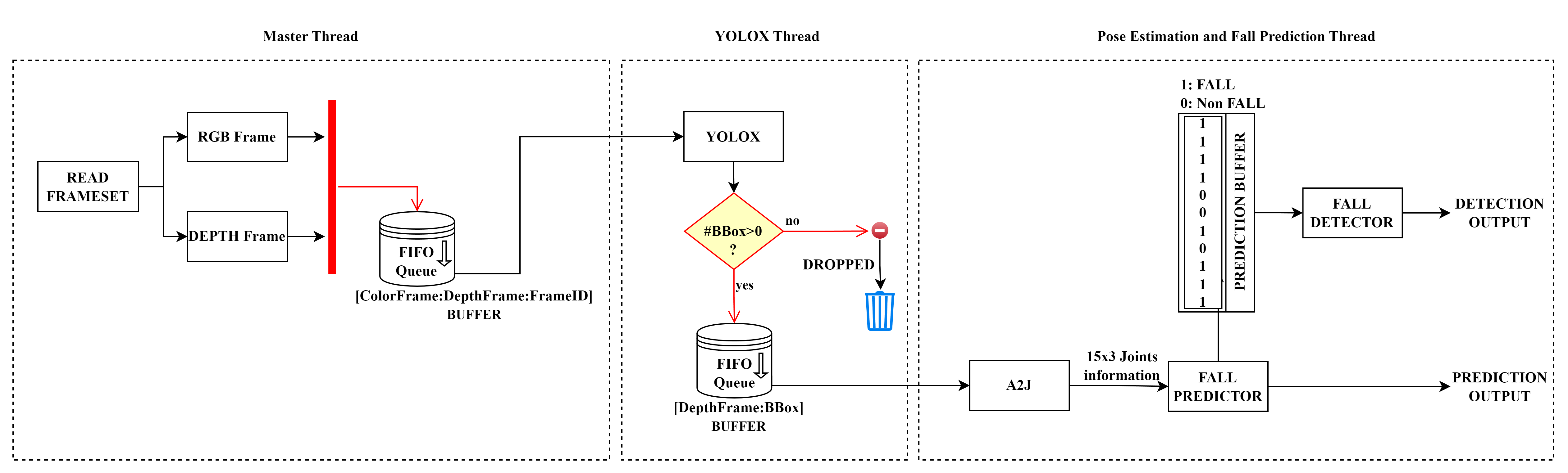}
    \caption{Parallel Design with Multi\-threaded Implementation.}
    \label{fig:mutlithreadedApplication}
\end{sidewaysfigure}
The parallel design implementation, shown in Figure \ref{fig:mutlithreadedApplication}, utilizes a multi-threaded approach and executes each AI model on a dedicated DPU core. A dual-core B1600 DPU architecture is synthesized, with YOLOX and A2J running on separate DPU cores, and the fall predictor model executing on the ARM CPU.

The main thread enqueues color and depth frames to a shared queue. The YOLOX thread continuously checks for frames in the queue, runs the YOLOX model on the color frame, and enqueues the depth frame and bounding box results for the pose estimation thread. The A2J thread dequeues depth frames, runs the pose estimation model, and passes the pose results to the fall prediction model for classification.

\subsection{Data Logging and Performance Monitoring} \label{sec:data-logging}
To monitor system performance and log relevant metrics, a signal handler is implemented to handle the \verb|SIGALRM| signal. The signal handler function is called periodically (every 100 milliseconds) by the system timer and logs various metrics to a CSV file, including timestamps, frame counts, frame rates, model inference times, and prediction outputs. This logging mechanism enables comprehensive analysis and evaluation of the system's performance.

%% file: 7_result.tex
\subsection{Training Results}
\begin{figure}[htbp]
    \centering
    \includegraphics[width=1\linewidth]{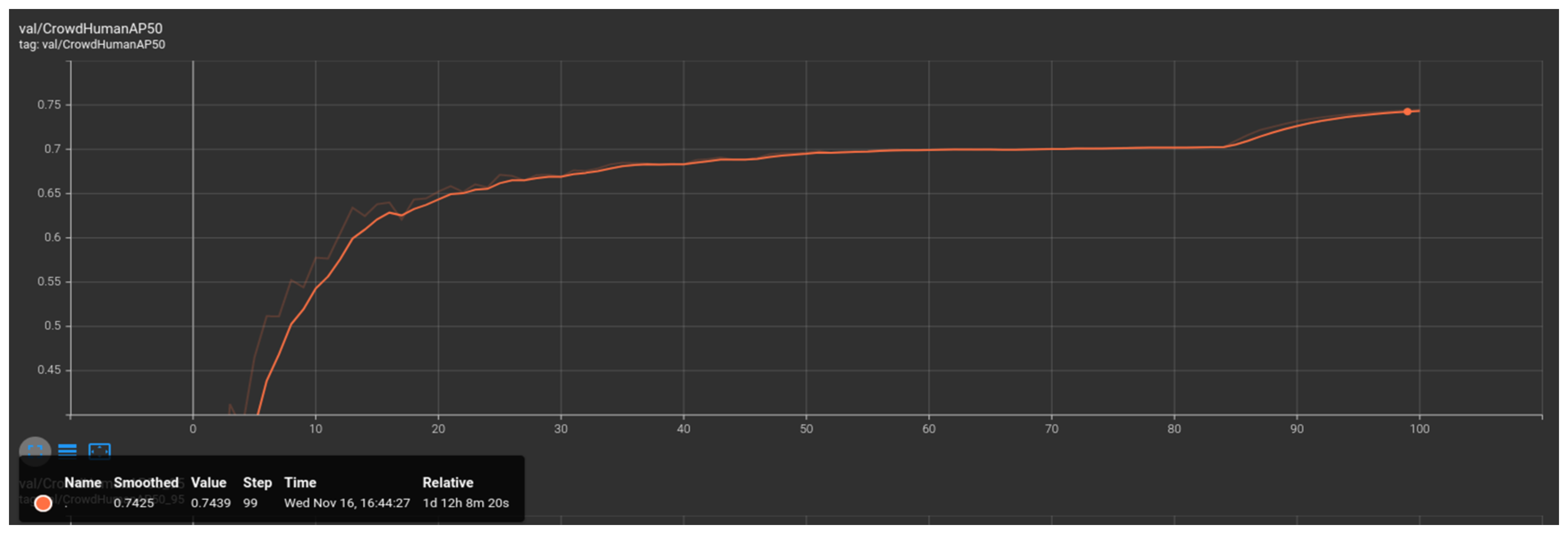}
    \caption{Training plot for the YOLOX object detection model, showing the IoU set at 50\% over the training epochs}
    \label{fig:yoloxTrainingPlot}
\end{figure}
The float model training results for the 3 models employed in the fall prediction and detection system are summarized here. Figure \ref{fig:yoloxTrainingPlot} illustrates the tensorboard training plot for the YOLOX object detection model, depicting the IoU set at 50\% over the training epochs. 
\begin{figure}[htbp]
    \centering
    \includegraphics[width=1\linewidth]{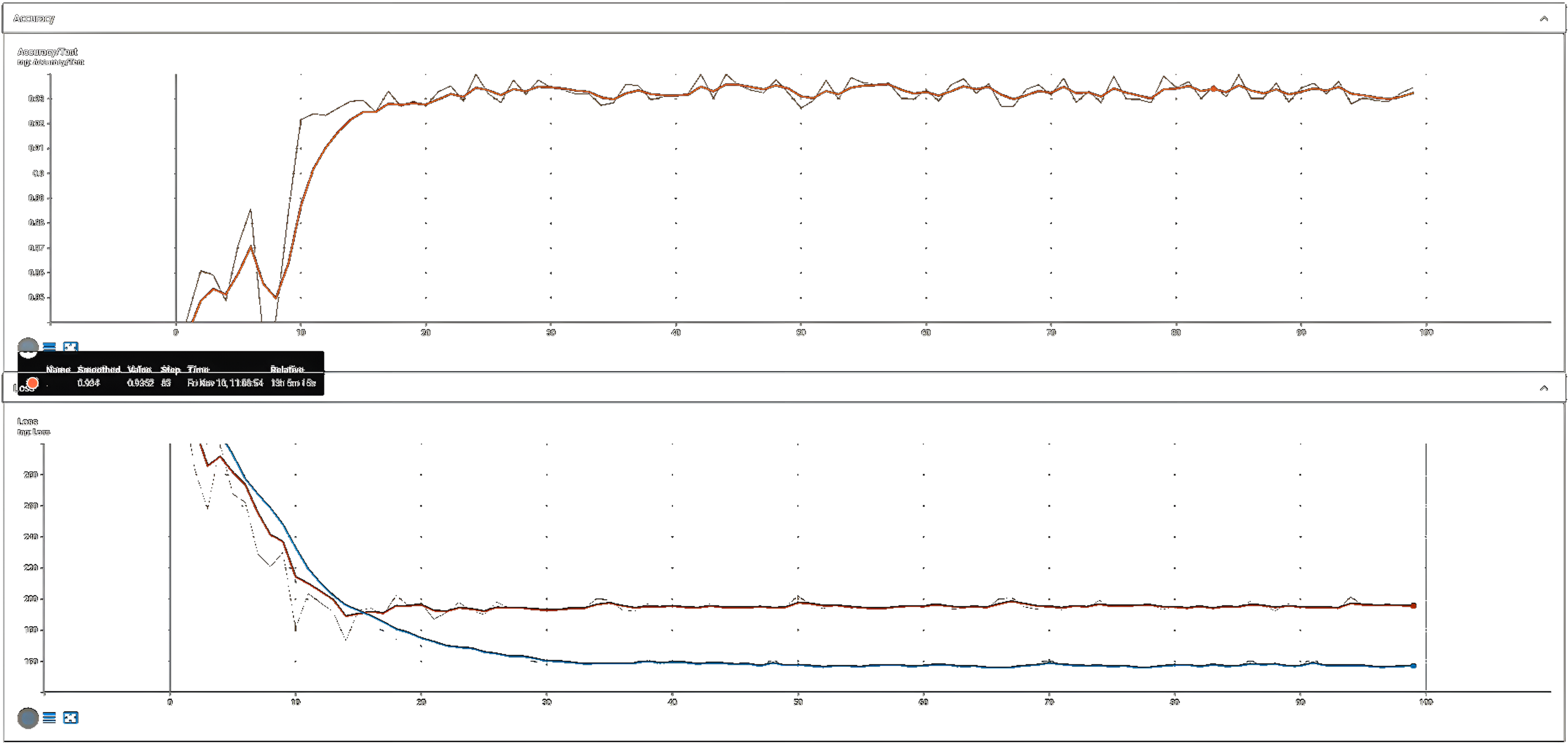}
    \caption{Training plot for the A2J human pose estimation model, top showing the accuracy of joint prediction, bottom plot shows the training vs validation loss.}
    \label{fig:a2jAccuracyPlot}
\end{figure}
Figure \ref{fig:a2jAccuracyPlot} shows the accuracy and training vs validation loss plot of the human pose estimation model A2J\cite{a2j}. \\
\begin{figure}[htbp]
    \centering
    \includegraphics[width=1\linewidth]{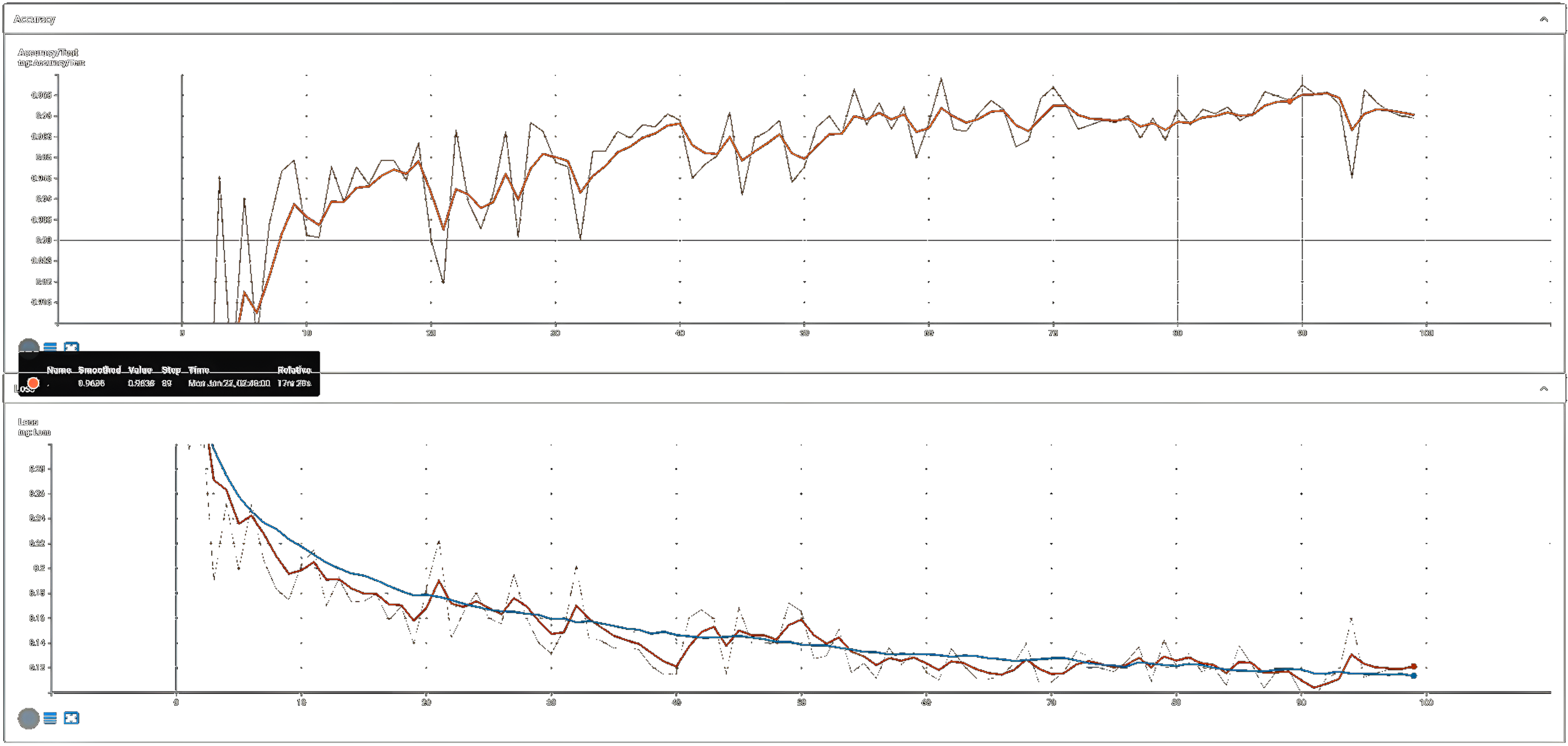}
    \caption{Training plot for the fall prediction model, top plot showing the accuracy, bottom plot shows the training vs validation loss.}
    \label{fig:fpTrainingPlot}
\end{figure}
Figure \ref{fig:fpTrainingPlot} presents the training plot for the fall prediction model, which utilizes temporal information and human pose data to anticipate potential falls before they occur. \\

\subsection{Quantization Results}
This section discusses the model performance once the quantized model is deployed on the hardware. Average power consumption by the SOM is monitored during the execution of the models, benchmarking tests are performed to identify the maximum throughput of a model executing instructions only on the DPU engine. 

\subsubsection{Quantized model summary}
The accuracy of the 3 model design is summarized in Table \ref{tab:model_performance_summary}. The FP32 is the float model accuracy that is trained on the GPU server. INT8 is the quantized model accuracy.
\begin{table}[htbp]
\resizebox{\linewidth}{!}{%
\centering
\begin{tabular}{|l|c|c|c|}
\hline
\textbf{Model} & \textbf{Dataset} & \textbf{Accuracy (FP32)} & \textbf{Accuracy (INT8)}\\ \hline
YOLOX$^{\mathrm{1}}$ & CrowdHuman & 74.45\% & 74.00\% \\ \hline
A2J$^{\mathrm{2}}$ & SDSU\_PSG & 93\% & 84.13\% \\ \hline
Fall Detector$^{\mathrm{3}}$ & SDSU\_PSG Fall Detection & 96.91\% & 75.85\% \\ \hline
\multicolumn{4}{l}{$^{\mathrm{1}}$Accuracy is based on the IoU of the boundary box, set at 50\%.} \\
\multicolumn{4}{p{82mm}}{$^{\mathrm{2}}$Accuracy is mean average precision (mAP) with a 10-cm rule for all the 15 joints.} \\
\multicolumn{4}{l}{$^{\mathrm{3}}$Accuracy = (TP + TN) / (TP + TN + FP + FN).} \\
\end{tabular}%
}
\caption{Model Performance Summary}
\label{tab:model_performance_summary}
\end{table}

\subsection{Power Efficiency}
To assess the power efficiency of the deployed system on the AMD Kria K26 SOM, the \verb|xmutil| command is used to monitor the power consumption of the platform. The following command output shows the real-time power consumption:
\begin{lstlisting}[style=command, language=bash]
ramesh@notos:~$ watch -n 1 xmutil xlnx_platformstats -p
\end{lstlisting}
The power consumption is higher in the parallel design implementation compared to the serial design implementation. In figure \ref{fig:benchmarkTable4096} and \ref{fig:benchmarkTable1600} first column shows the power consumption in milliwatts (mW) for individual models executing.

\subsection{Benchmark results}
\begin{figure}[htbp]
    \centering
    \includegraphics[width=1\linewidth]{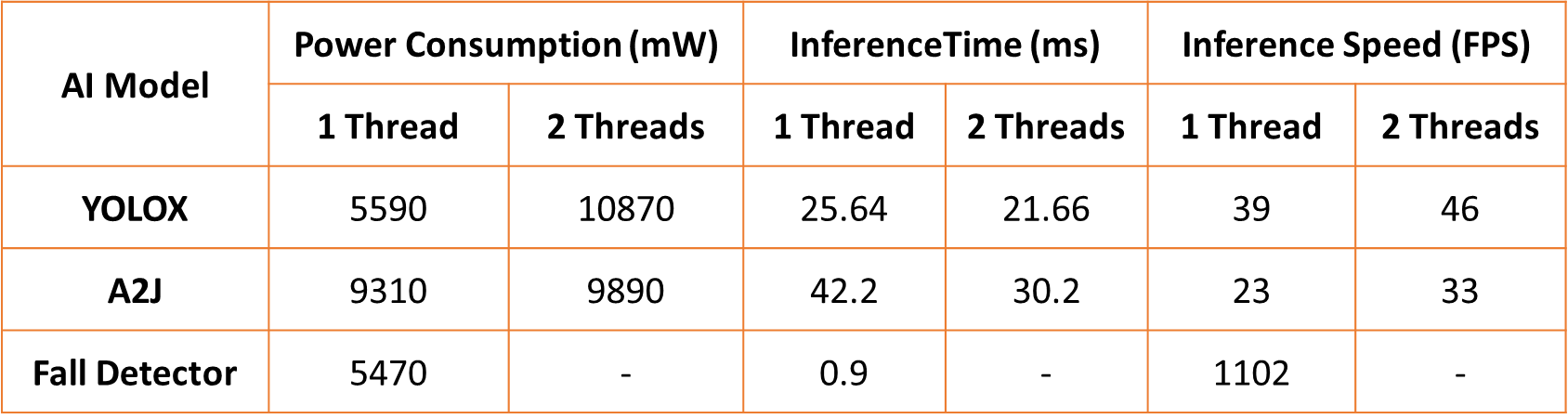}
    \footnotemark{Does not include the frame capture time (only model inference on the DPU without pre/post processing).}
    \caption{Benchmark Performance of AI Models on B4096 architecture DPU.}
    \label{fig:benchmarkTable4096}
\end{figure}

\begin{figure}[htbp]
    \centering
    \includegraphics[width=1\linewidth]{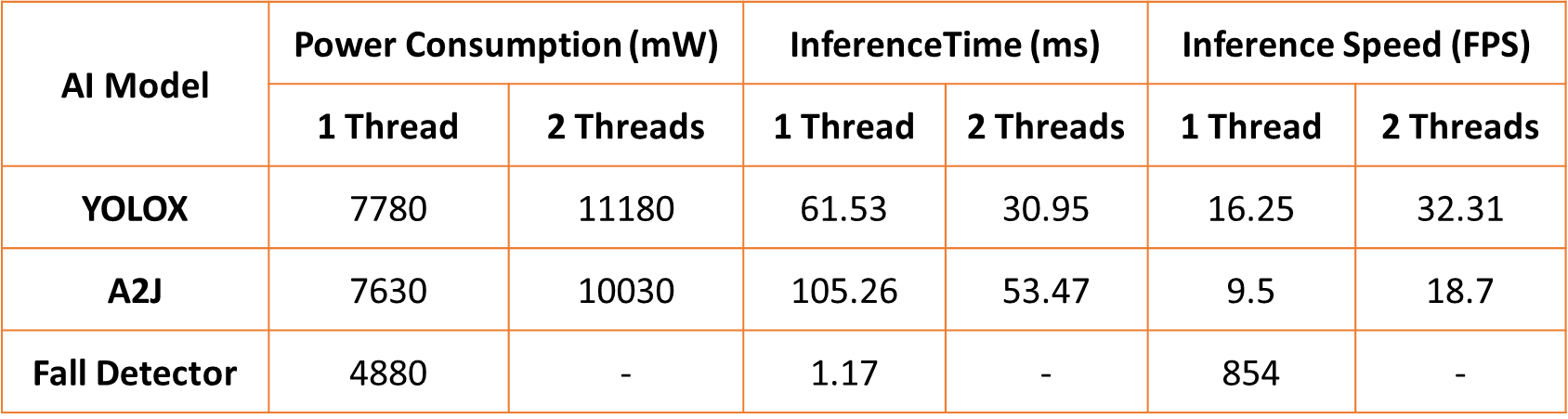}
    \footnotemark{Does not include the frame capture time (only model inference on the DPU without pre/post processing).}
    \caption{Benchmark Performance of AI Models on B1600 architecture DPU.}
    \label{fig:benchmarkTable1600}
\end{figure}

The \verb|xdputil| command is used to benchmark the performance of the quantized models on the AMD Kria K26 SOM. The output is given in the units of frames per second (FPS).
The inference time in milliseconds, inference speed in FPS, and power consumption in milliwatts (mW) is summarized for each models in figure \ref{fig:benchmarkTable4096} and figure \ref{fig:benchmarkTable1600} running on B4096 and B1600 architecture DPUs, respectively.

\subsection{Profiling Results: } \label{prfilingResults}
To analyze the performance and identify potential bottlenecks, profiling was conducted on the deployed system using the AMD Kria K26 SOM.

\subsubsection{Frame Alignment Bottleneck}
During the profiling process of the producer process that simply reads the frame and flushes it to consumer, a bottleneck was identified in the frame alignment step, which aligns the depth and color frames for accurate boundary box estimation. The following terminal output illustrate the time taken by the \verb|align_to_depth| function:

\begin{lstlisting}[style=command, language=bash]
root@xilinx-kv260-starterkit-20221:~/camera# ./profiling
align_to_depth Elapsed time: 115 ms
align_to_depth Elapsed time: 91 ms
align_to_depth Elapsed time: 94 ms
align_to_depth Elapsed time: 88 ms
align_to_depth Elapsed time: 87 ms
align_to_depth Elapsed time: 90 ms
\end{lstlisting}

\subsubsection{Vitis AI Analyzer}
\begin{figure}[htbp]
    \centering
    \includegraphics[width=1\linewidth]{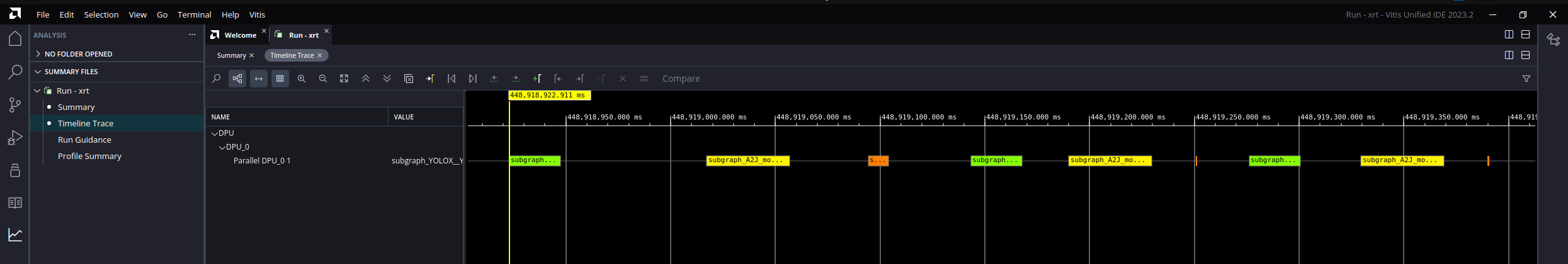}
    \caption{Vitis AI Analyzer view showing profile results of 1 core DPU.}
    \label{fig:profile_1core}
\end{figure}

\begin{figure}[htbp]
    \centering
    \includegraphics[width=1\linewidth]{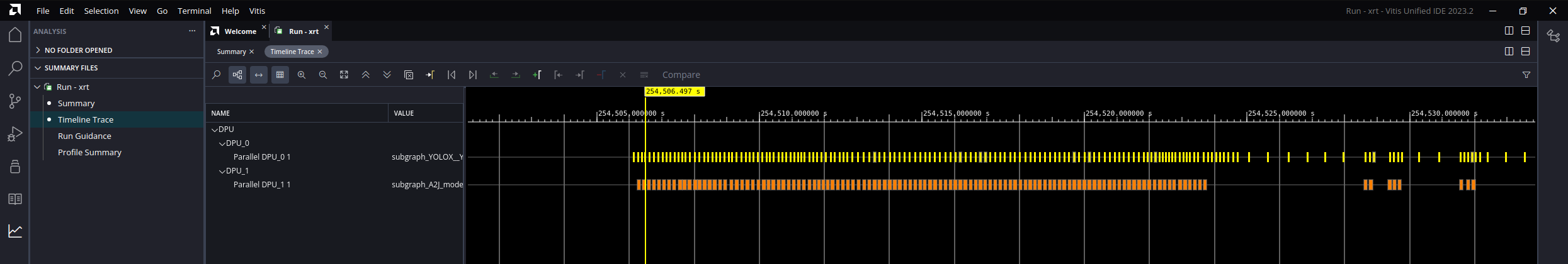}
    \caption{Vitis AI Analyzer view showing profile results of 2 core DPU.}
    \label{fig:profile_2core}
\end{figure}

The Vitis AI Analyzer is used to profile the performance of the deployed models on the AMD Kria K26 SOM's DPU. Figures \ref{fig:profile_1core} and \ref{fig:profile_2core} show the Vitis AI Analyzer views for profiling results with one core and two cores DPU, respectively.
In figure \ref{fig:profile_1core} we observe the time between the two subgraphs execution on the DPU is the time spent on CPU executing the post processing and preprocessing functions. This is an area to improve the system performance and the overall throughput.
Figure \ref{fig:profile_2core} shows the execution of the DPU subgraphs executing on the two cores of DPU. Each model is dedicated to a DPU core for its execution. 
\begin{figure}[htbp]
    \centering
    \includegraphics[width=1\linewidth]{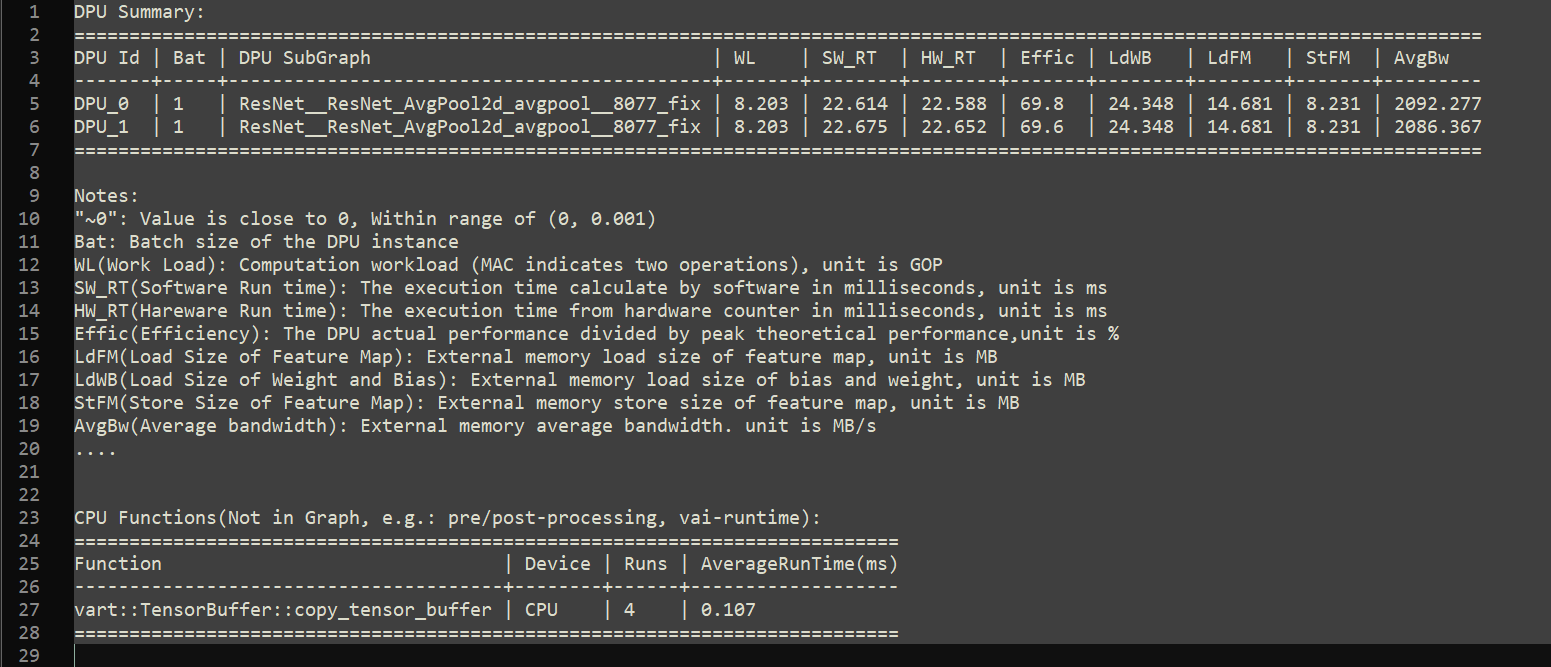}
    \caption{Text Summary of the DPU profile.}
    \label{fig:txtSummary}
\end{figure}
Figure \ref{fig:txtSummary} shows an example of text summary that is printed on the terminal when the profiling is executed. The figure shows important parameters such as a core usage efficiency of 69.8 and 69.6 on DPU core one and core two, respectively, external memory average bandwidth in MBps, and computation work-load on each core that indicates the number of MAC operations in GOP units.

\subsection{Serial Implementation}
\begin{figure}[htbp]
    \centering
    \includegraphics[width=1\linewidth]{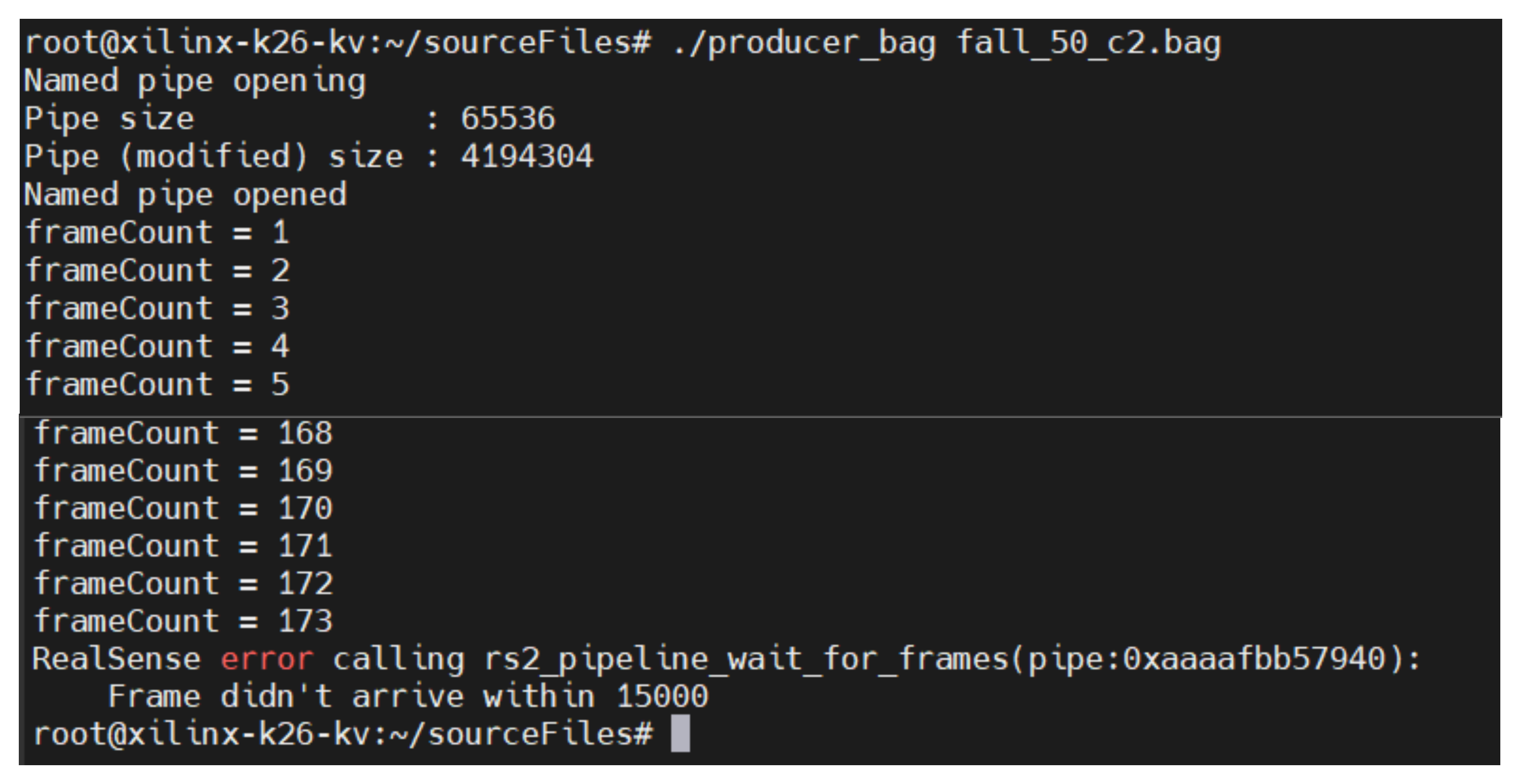}
    \caption{Log showing the execution of the producer process.}
    \label{fig:producerTerminalLog}
\end{figure}

The first step is to start the execution of the application is to run the producer process. Figure \ref{fig:producerTerminalLog} shows the execution of the producer process that takes a \verb|.bag| file as its input. The producer process prints the frame number and other exceptions during runtime.
\begin{figure}[htbp]
    \centering
    \includegraphics[width=1\linewidth]{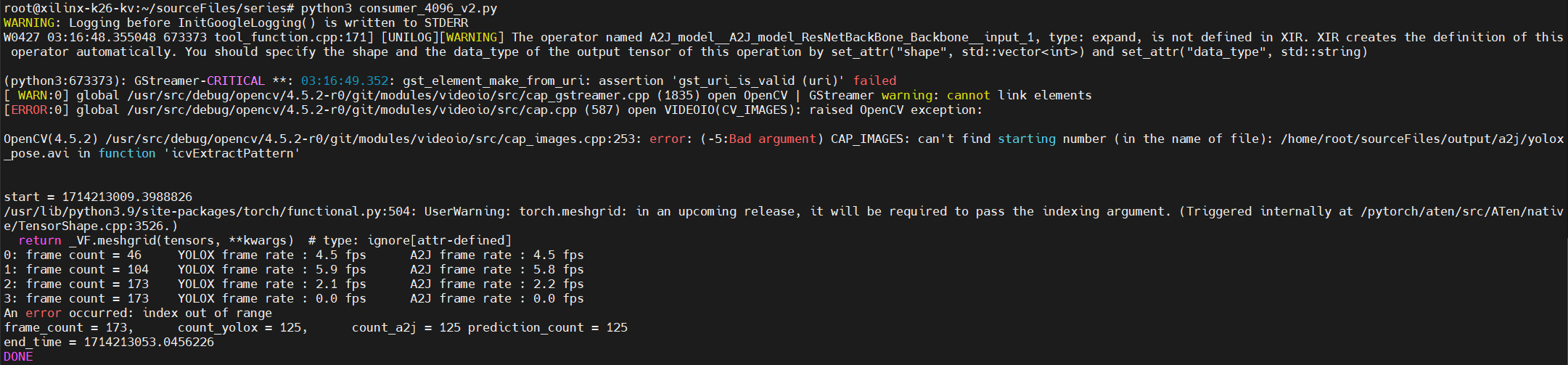}
    \caption{Log showing the execution of the consumer process with serial pipeline implementation}
    \label{fig:consumerSerialLog}
\end{figure}

\noindent\textbf{Summary of the serial design implementation:}
\begin{enumerate}
    \item The bag file used to run the test have 175 frames that was recorded at 30 FPS.
    \item Figure \ref{fig:consumerSerialLog} shows the terminal log when the consumer application is executed. 
    \item Yolox model was able to infer 125 frames out of 175 frames.
    \item All the 125 frames were processed by the A2J model and generated the pose.
    \item 125 predictions were made on the pose data.
    \item The total run time of the consumer application = 43.65 seconds.
\end{enumerate}
The consumer application implements a log of the number of frames processed by each model and its time stamps. 
\begin{figure}[htbp]
    \centering
    \includegraphics[width=1\linewidth]{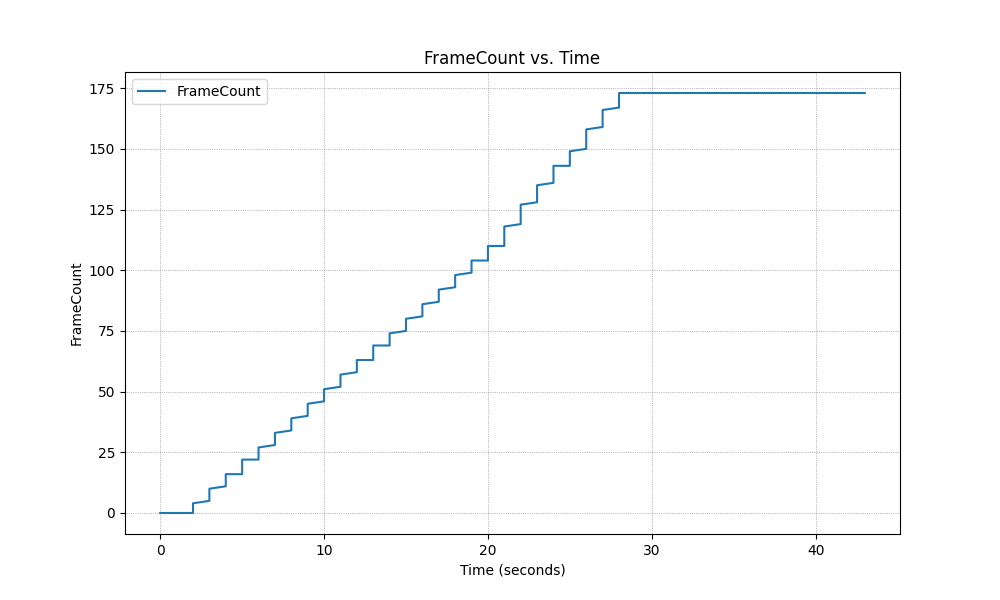}
    \caption{Number of frames received vs. relative time in the serial implementation.}
    \label{fig:1_FrameCount_vs_Time_serial}
\end{figure}
Figure \ref{fig:1_FrameCount_vs_Time_serial} shows the total number of frames seen by the consumer application vs relative time.
\begin{figure}[htbp]
    \centering
    \includegraphics[width=1\linewidth]{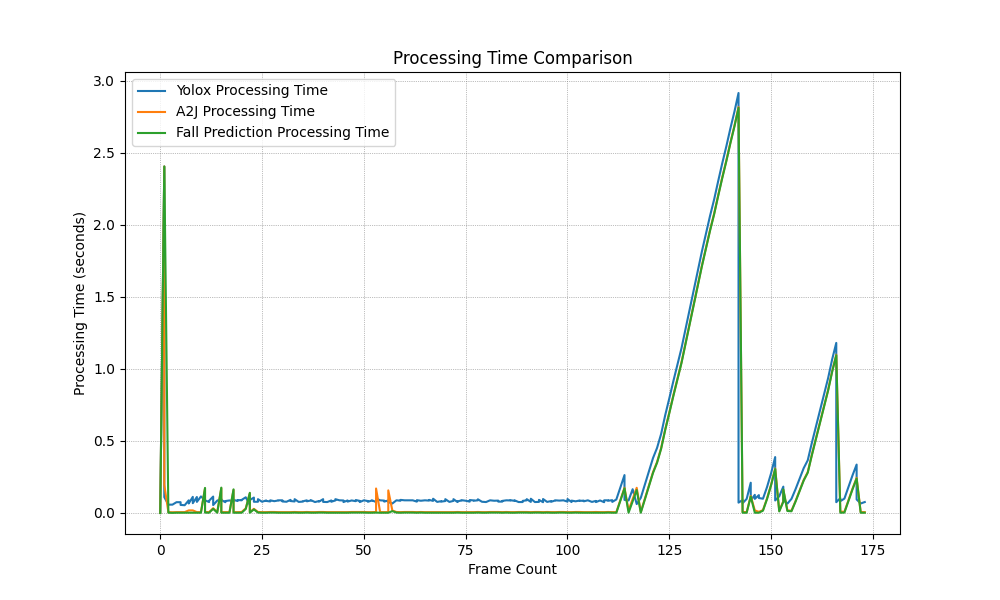}
    \caption{Time difference between the frame arrival time and inference time by each model in the serial implementation (includes all CPU pre/post processing).}
    \label{fig:4_Processing_Time_Comparison_serial}
\end{figure}
Figure \ref{fig:4_Processing_Time_Comparison_serial} shows the difference in time between the frame originally arrived to the consumer and the time taken by each model for inferencing, including the pre- and post-processing functions. The large spike observed for the frames towards the ending of the video (frames that featured a person falling or with bad posture) shows that the YOLOX was unable to infer the person. A significant number of frames were skipped due to this problem.
\begin{figure}[htbp]
    \centering
    \includegraphics[width=1\linewidth]{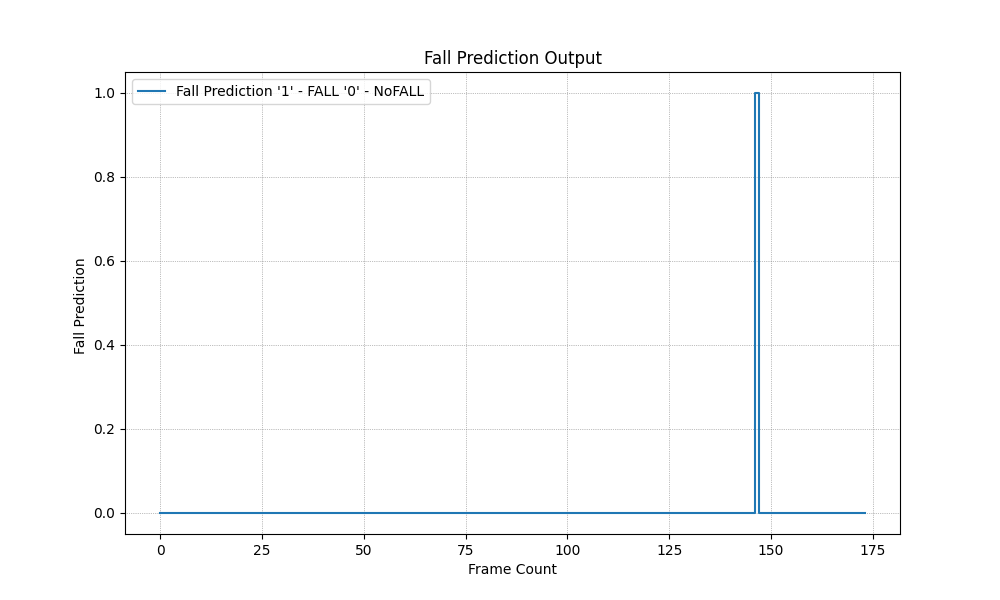}
    \caption{Frames predicted as FALL/No FALL in Serial Implementation.}
    \label{fig:3_Fall_Prediction_Output_serial}
\end{figure}
Figure \ref{fig:3_Fall_Prediction_Output_serial} shows the final prediction output for each frames. A \textit{FALL} is represented with '1' and a \textit{No Fall} as '0'. The prediction is a no-fall after the fall has occured could be due to multiple reasons, One of which could be false positives prediction as an ADL activity of sleeping.  

\subsection{Parallel Implementation}
The first step is to start the execution of the application is to run the producer process. \\
\begin{figure}[htbp]
    \centering
    \includegraphics[width=1\linewidth]{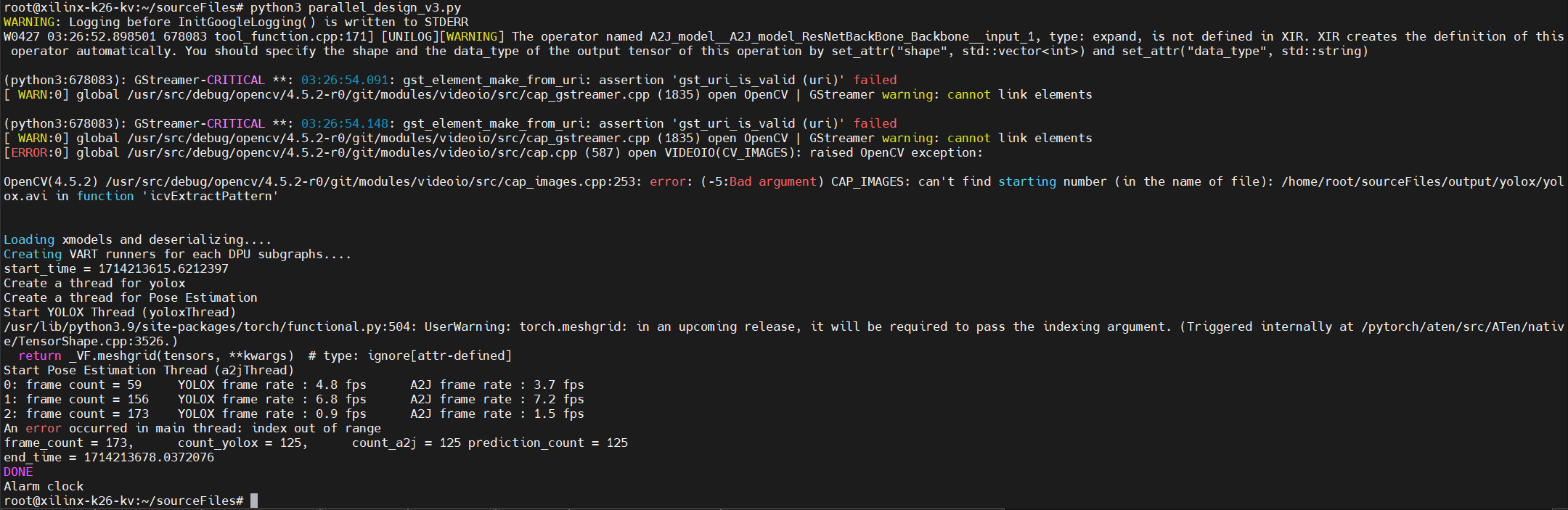}
    \caption{Log showing the execution of the consumer process with serial pipeline implementation}
    \label{fig:consumerParallelLog}
\end{figure}
\noindent\textbf{Summary of the parallel design implementation:}
\begin{enumerate}
    \item The bag file used to run the test have 175 frames that was recorded at 30 FPS.
    \item Figure \ref{fig:consumerParallelLog} shows the terminal log when the consumer application is executed. 
    \item Yolox model was able to infer 125 frames out of 175 frames.
    \item All the 125 frames were processed by the A2J model and generated the pose. 
    \item 125 predictions were made on the pose data.
    \item The total run time of the consumer application = 62.42 seconds.
    \item The difference in the serial and parallel design implementation is that the throughput of individual model is improved.
\end{enumerate}
\begin{figure}[htbp]
    \centering
    \includegraphics[width=1\linewidth]{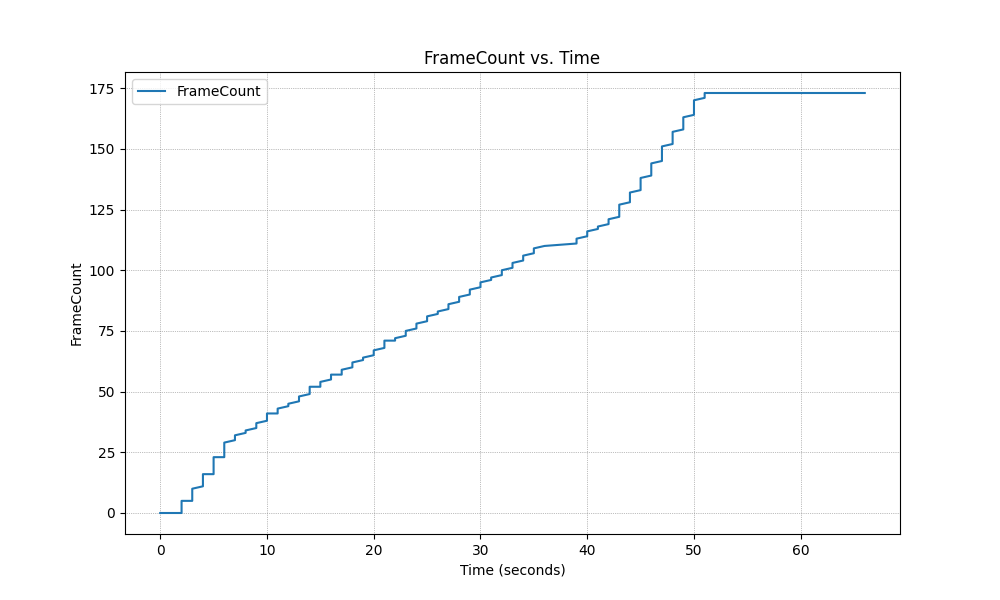}
    \caption{Number of frames received vs relative time in Parallel Implementation.}
    \label{fig:1_FrameCount_vs_Time_parallel}
\end{figure}
Figure \ref{fig:1_FrameCount_vs_Time_parallel} shows the total number of frames seen by the consumer application vs relative time.
The fall prediction model executes as soon as the A2J model produces the pose joint information. We can see that YOLOX model continues its execution until there are new frames available. Similarly, the A2J and Fall predictor models execute as soon as the YOLOX model produces a valid frame. 
\begin{figure}[htbp]
    \centering
    \includegraphics[width=1\linewidth]{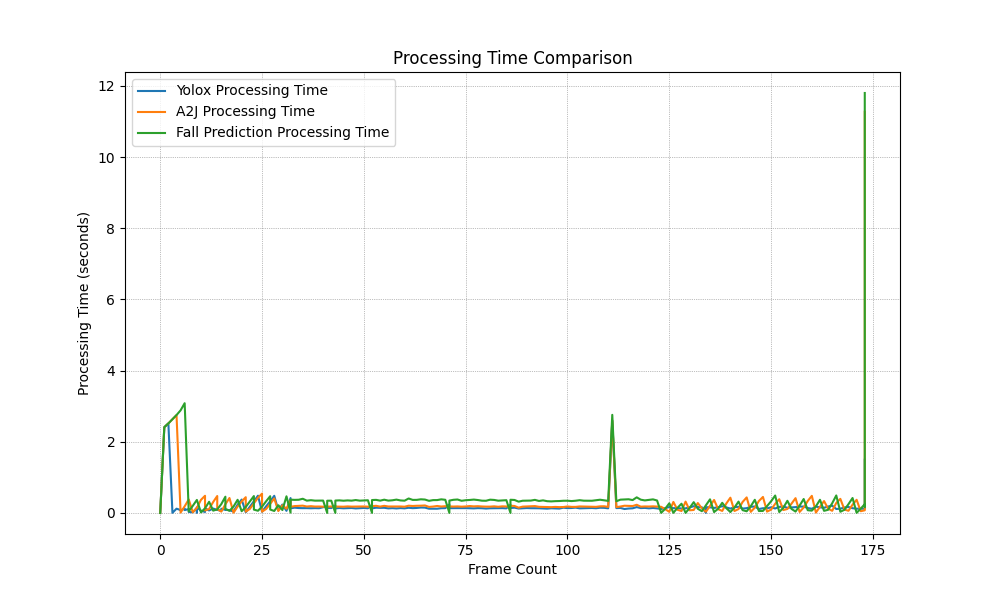}
    \caption{Time difference between the frame arrival time and inference time by each model in the parallel implementation (includes all CPU pre/post processing).}
    \label{fig:4_Processing_Time_Comparison_parallel}
\end{figure}
Figure \ref{fig:4_Processing_Time_Comparison_parallel} shows the difference in time between a frame that originally arrives to the consumer frame queue and the time taken by each model for inferencing, including the pre- and post-processing functions. The large spike observed for the last frame is due to the maximum wait time of the frame from the producer showing \textit{end of frame} in the video file.
\begin{figure}[htbp]
    \centering
    \includegraphics[width=1\linewidth]{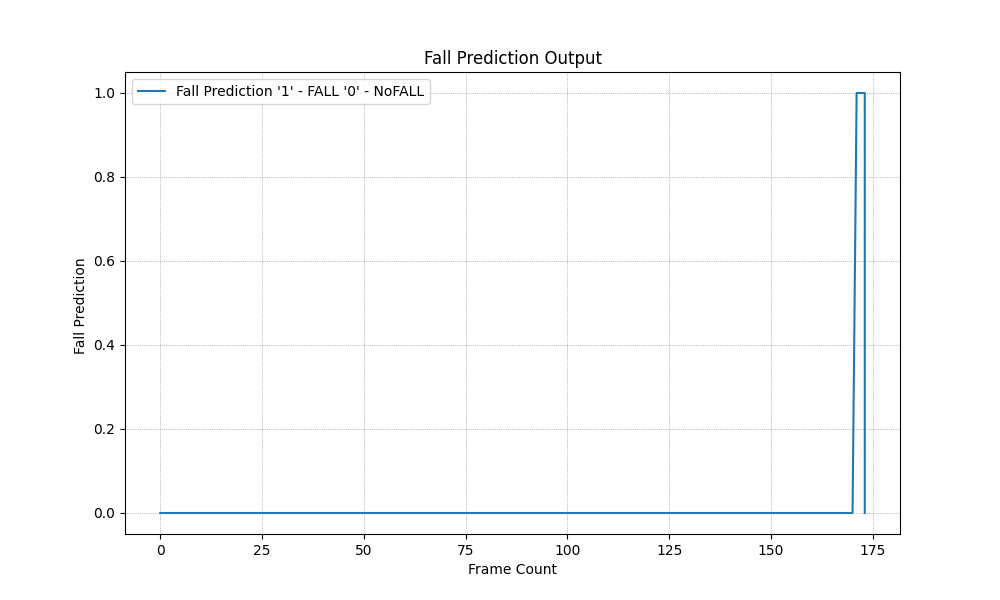}
    \caption{Frames predicted as FALL/No FALL in the parallel implementation.}
    \label{fig:3_Fall_Prediction_Output_parallel}
\end{figure}
Figure \ref{fig:3_Fall_Prediction_Output_parallel} shows the final prediction output for each frames. FALL is represented with '1' and a No Fall as '0'.

\subsection{System Throughput}
\begin{figure}[htbp]
    \centering
    \includegraphics[width=1\linewidth]{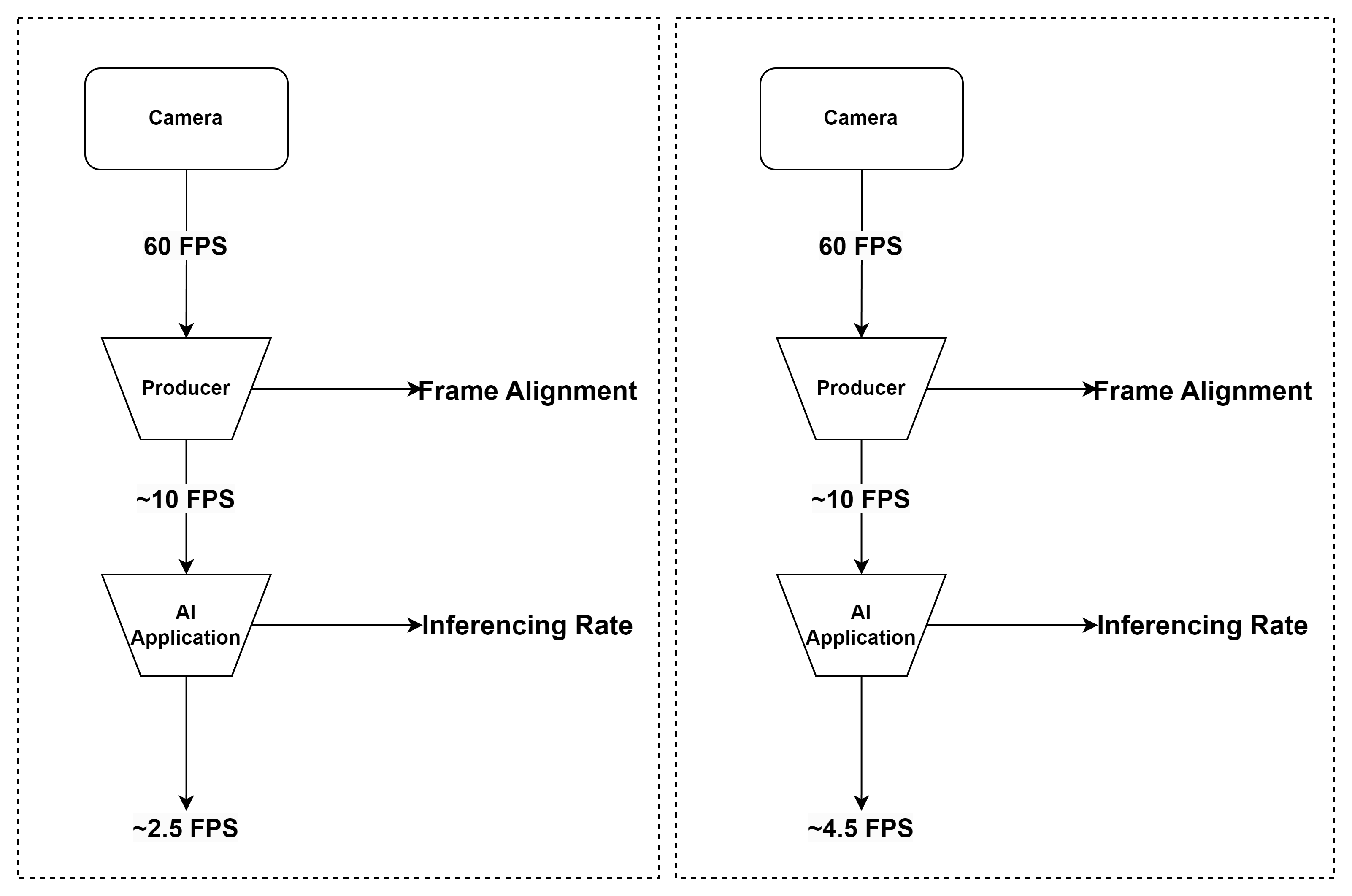}
    \caption{Bottleneck diagram showing the throughput rate for serial (left) and parallel (right) design implementation.}
    \label{fig:bottleneck}
\end{figure}
Figure \ref{fig:bottleneck} shows the system throughput analysis of the deployed system, highlighting the overall performance and capabilities for the serial and parallel design implementation. The first bottleneck in the application is the frame alignment process which is a computationally costly operation on the CPU. The inferencing rate is improved in the parallel design implementation for individual models. Therefore, the final throughput is comparatively higher than the serial implementation.

\subsection{Fall Prediction Results}
\label{Fall Prediction Results}
\begin{figure}[htbp]
    \centering
    \includegraphics[width=1\linewidth]{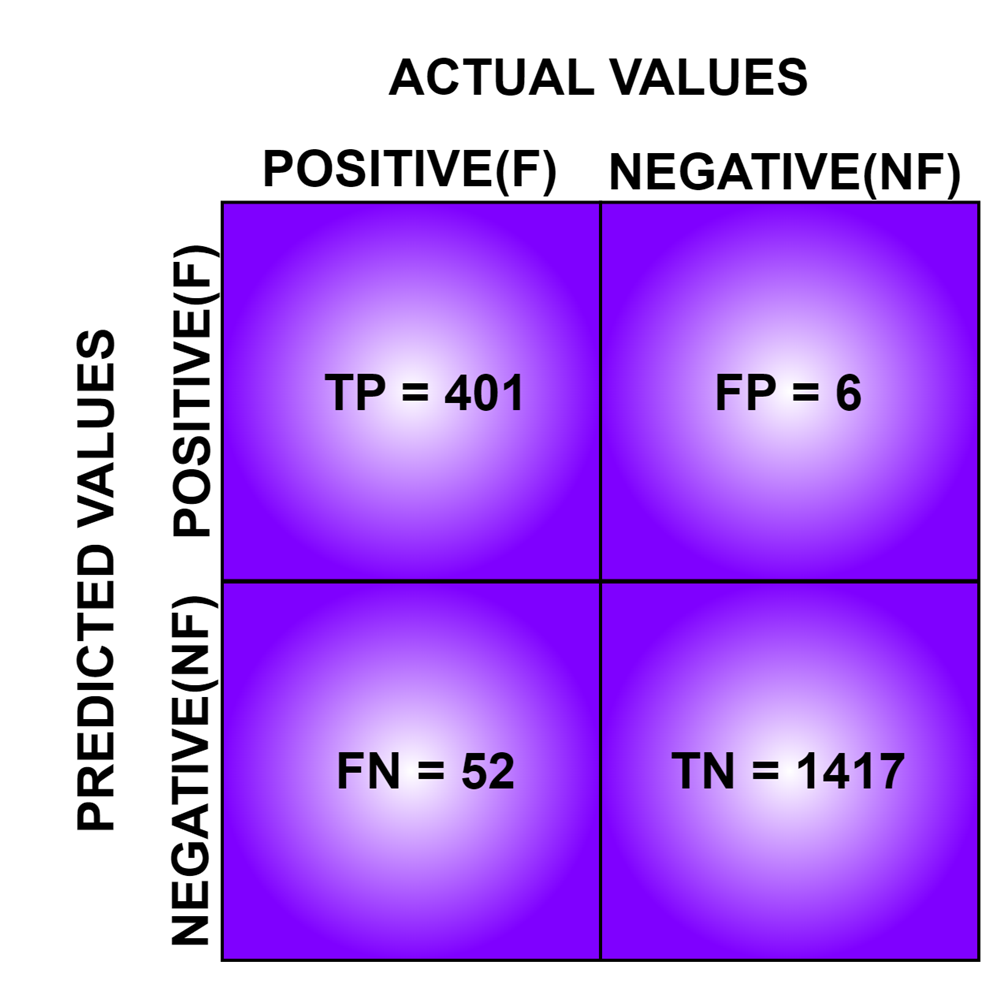}
    \caption{Confusion matrix of the fall predictor model.}
    \label{fig:confusionMatrix}
\end{figure}
The performance of the proposed system in predicting falls was evaluated using the validation set of the \verb|SDSU_PSG| dataset, and the results are presented in this section. The system's prediction capabilities were assessed through various metrics, including a confusion matrix (Figure \ref{fig:confusionMatrix}, F1 score, precision, recall, and overall accuracy. The confusion matrix provides insights into the system's ability to correctly classify falls and non-falls, while the F1 score, precision, and recall metrics quantify the trade-off between false positives and false negatives. The achieved accuracy score highlights the system's overall performance in accurately predicting falls and non-falls.
The number of frames involving Fall are 453 and non-fall are 1423.
\begin{equation}
    \text{F1-score} = 2 \times \frac{\text{precision} \times \text{recall}}{\text{precision} + \text{recall}}
\end{equation}

\begin{equation}
    \text{Precision} = \frac{\text{True Positives}}{\text{True Positives} + \text{False Positives}}
\end{equation}

\begin{equation}
    \text{Recall} = \frac{\text{True Positives}}{\text{True Positives} + \text{False Negatives}}
\end{equation}

\begin{equation}
    \text{Accuracy} = \frac{\text{TP} + \text{TN}}{\text{TP} + \text{FP} + \text{TN} + \text{FN}}
\end{equation}
\begin{equation}    
   F_\beta = (1 + \beta^2) \cdot \frac{\text{Precision} \cdot \text{Recall}}{(\beta^2 \cdot \text{Precision}) + \text{Recall}}
\end{equation}
Where,\\ 
True Positives (TP) are the instances correctly predicted as FALL.\\
True Negatives (TN) are the instances correctly predicted as NOFALL.\\
False Positives (FP) are the instances incorrectly predicted as FALL.\\
False Negatives (FN) are the instances incorrectly predicted as NOFALL.\\
The values obtained from the confusion matrix are: \\
\begin{align*}
    F1\ Score &:\ 0.93\\
    Precision &:\ 0.99\\
    Recall &:\ 0.89\\
    Accuracy &:\ 0.97\\
    \textcolor{blue}{F3\ Score }&:\ \textcolor{blue}{0.89}
\end{align*}

\subsection{Sample Inference Results}
\begin{sidewaysfigure}
    \centering\includegraphics[width=1\linewidth]{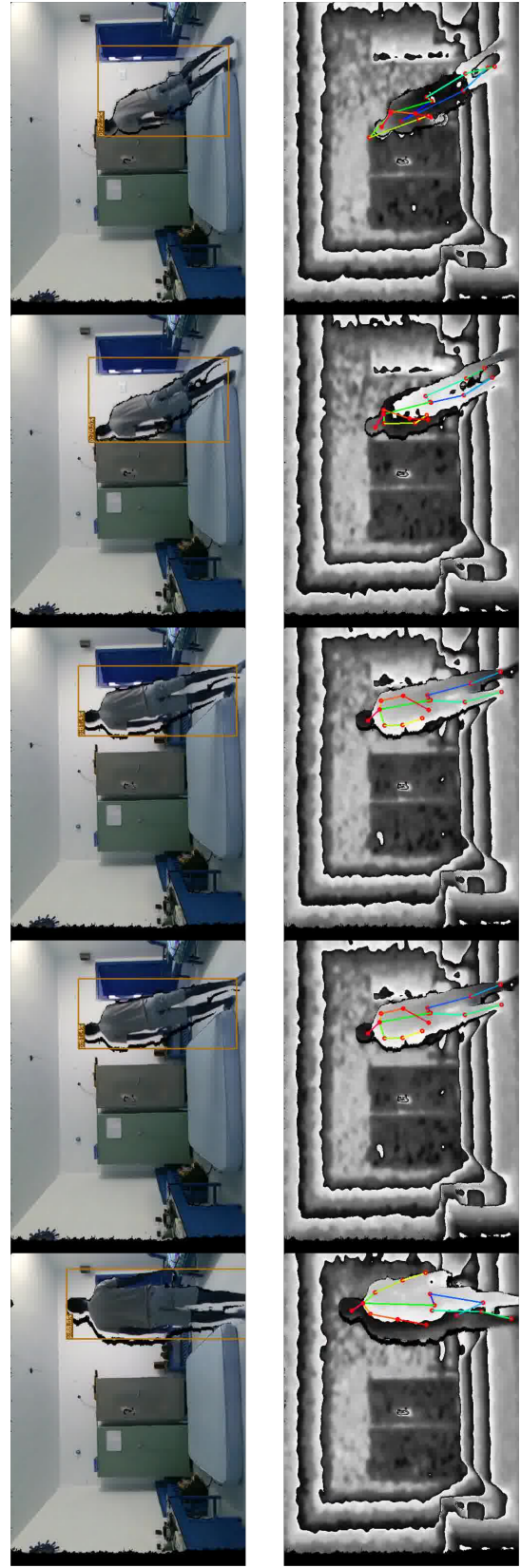}
    \caption{Output captured at intermediate stages of the pipeline.}
    \label{fig:finalOutputImages}
\end{sidewaysfigure}
Figure \ref{fig:finalOutputImages} shows the output of frames captured at intermediate stages of the pipeline.

%% file: 9_discussion.tex
The development of edge-based fall detection systems presents unique challenges in balancing performance, privacy, and practical implementation. Our research findings provide several insights into these challenges while suggesting potential pathways for future development.

\subsection{Edge Computing Performance}
Our implementation on the K26 SOM demonstrates the viability of edge-based fall detection, achieving ~5.5 FPS with multi-threading optimization. This aligns with the current trend toward edge computing solutions identified by Mrozek et al. \cite{Grzesik2024}, though with some performance trade-offs. While traditional fall detection systems often rely on cloud computing for processing \cite{El_Attaoui2020}, our edge-based approach eliminates latency issues and privacy concerns associated with data transmission, albeit at the cost of lower frame rates.

\subsection{Privacy-Preserving Architecture}
The system's architecture addresses a critical gap in existing fall detection solutions by incorporating privacy preservation at its core. By discarding RGB frames after initial processing and relying on depth data for analysis, our approach aligns with privacy-preserving techniques recommended by Liu et al. \cite{Liu2020}. This design choice is particularly relevant for deployment in sensitive environments such as healthcare facilities and private residences.

\subsection{Model Performance and Trade-offs}
The accuracy rates achieved by our pipeline (YOLOX: 74\%, A2J: 84.13\%, Fall prediction: 75.85\%) reveal interesting insights about model performance on edge devices. While these figures are lower than some cloud-based solutions \cite{Huang2018}, they represent a practical compromise between accuracy and real-time processing capabilities. The relatively high accuracy of the A2J model (84.13\%) suggests that joint detection remains robust even with edge computing constraints.

\subsection{System Limitations and Practical Considerations}
Several key limitations warrant discussion:
\begin{enumerate}
    \item The frame rate limitation may impact the system's ability to detect rapid falls
    \item Current accuracy levels, while promising, may need improvement for clinical applications
    \item Environmental factors such as occlusions and multiple subjects may affect performance
\end{enumerate}
These limitations align with challenges identified in previous research \cite{Zhang2015}, \cite{Igual2013}, suggesting common hurdles in edge-based vision systems.

\subsection{Future Development Paths}
This research opens several avenues for future investigation:
\begin{enumerate}
    \item Implementing lightweight model architectures to improve frame rates while maintaining accuracy
    \item Exploring hardware acceleration techniques for enhanced real-time performance
    \item Conducting extensive real-world testing in elderly care environments
    \item Investigating methods to improve model accuracy without compromising privacy or computational efficiency
\end{enumerate}

%% file: 8_conclusion_and_future_works.tex
In this work, we presented a comprehensive vision-based fall prediction and detection system leveraging human pose estimation from depth camera video on the AMD Kria K26 SOM platform. The proposed method meets the design goals identified in Section \ref{introduction} and is a novel approach to fall detection representing a promising step towards enhancing the safety, independence, and overall quality of life for vulnerable populations, while reducing the burden on healthcare systems and caregivers. Through continued research and innovation, the developed system can be further refined and optimized, ultimately enabling widespread deployment and making a tangible impact on the lives of those at risk of falls.